\documentclass{article}
\usepackage[final]{colm2026_conference}

\usepackage{hyperref}
\usepackage{url}
\usepackage{booktabs}
\usepackage{lineno}
\usepackage{color}
\usepackage{tcolorbox}
\usepackage{caption}
\tcbuselibrary{breakable, skins}

\newtcolorbox{promptbox}[1]{
    breakable,
    colback=white,
    colframe=black!70,
    fonttitle=\scriptsize\bfseries,
    title=#1,
    toprule=1.5pt,
    bottomrule=0.5pt,
    left=3pt, right=3pt, top=3pt, bottom=3pt,
}
\newtcolorbox{systembox}{
    breakable,
    colback=white,
    colframe=black!40,
    leftrule=3pt,
    rightrule=0.4pt,
    toprule=0.4pt,
    bottomrule=0.4pt,
    fontupper=\scriptsize,
    title={\scriptsize\textbf{System}},
    left=3pt, right=3pt, top=2pt, bottom=2pt,
    before skip=2pt, after skip=2pt,
}
\newtcolorbox{userbox}{
    breakable,
    colback=white,
    colframe=black!20,
    leftrule=3pt,
    rightrule=0.4pt,
    toprule=0.4pt,
    bottomrule=0.4pt,
    fontupper=\scriptsize,
    title={\scriptsize\textbf{User}},
    left=3pt, right=3pt, top=2pt, bottom=2pt,
    before skip=2pt, after skip=2pt,
}

\definecolor{ForestGreen}{rgb}{0.133,0.545,0.133}
\definecolor{BrickRed}{rgb}{0.796,0.255,0.329}
\definecolor{darkblue}{rgb}{0,0,0.5}
\definecolor{mygreen}{rgb}{0.133,0.545,0.133}

\hypersetup{
  colorlinks=true,
  citecolor=darkblue,
  linkcolor=darkblue,
  urlcolor=darkblue
}

\usepackage{amsmath,amsfonts,bm}

\def\eqref#1{equation~\ref{#1}}

\def\1{\bm{1}}

\DeclareMathAlphabet{\mathsfit}{\encodingdefault}{\sfdefault}{m}{sl}
\SetMathAlphabet{\mathsfit}{bold}{\encodingdefault}{\sfdefault}{bx}{n}

\usepackage{xcolor}
\usepackage{graphicx}
\usepackage{tabularx}
\usepackage{amsmath}
\usepackage[capitalize]{cleveref}
\usepackage{mathtools}
\usepackage{multirow}
\usepackage{setspace}
\hypersetup{colorlinks,linkcolor={blue},citecolor={blue},urlcolor={blue}}

\usepackage{latexsym}
\usepackage{comment}
\usepackage{tabulary}
\usepackage{amsfonts}
\usepackage{array}

\usepackage[T1]{fontenc}
\usepackage[utf8]{inputenc}
\usepackage{listings}
\usepackage{pifont}
\usepackage{inconsolata}

\usepackage{colortbl}
\usepackage{enumitem}
\usepackage{makecell}
\usepackage{arydshln}

\usepackage{subcaption}
\usepackage{wrapfig}
\usepackage{float}

\newcommand{\sig}{\textsuperscript{\tiny{*}}}
\newcommand{\algname}{\texttt{Q-CARE}}

\definecolor{absgray}{RGB}{242,243,245}
\definecolor{metablue}{RGB}{0,102,204}
\definecolor{lightred}{RGB}{255, 204, 204}
\definecolor{lightgreen}{RGB}{204, 255, 204}
\definecolor{lightyellow}{RGB}{255, 255, 204}

\newcommand{\customabstractpage}{
\begin{tcolorbox}[
    enhanced,
    colback=absgray,
    colframe=absgray,
    boxrule=0pt,
    arc=8pt,
    left=3mm,
    right=3mm,
    top=3mm,
    bottom=3mm
]

{\Large\bfseries
Towards Query-Agnostic RAG Evaluation via Query Coverage and Claim Verifiability
\par}

\vspace{3mm}

\textbf{Jeonghwan Choi}$^{1}$, \textbf{Taewon Yun}$^{1}$, \textbf{Minjeong Ban}$^{1}$, \textbf{Gyeonghun Sun}$^{1}$\\ \textbf{Jae-Gil Lee}$^{1}$, \textbf{Hwanjun Song}$^{1,2,*}$\par

\vspace{1mm}

$^{1}$Korea Advanced Institute of Science and Technology (KAIST), $^{2}$Cluvion\\\par

\vspace{4mm}

\noindent
Retrieval-augmented generation improves the factuality of large language models by grounding responses in retrieved evidence, yet existing evaluation frameworks struggle to provide consistent, fine-grained diagnostics across the diverse spectrum of user queries, ranging from close-ended fact-seeking to open-ended explanatory requests. We propose \algname{}, a query-agnostic and fully reference-free framework that enables fine-grained assessment by decomposing queries into sub-queries and answers into atomic claims. \algname{} establishes a unified evaluation principle based on query coverage and claim verifiability, yielding coverage-aware retriever metrics (C-Prec@k, C-nDCG@k) and claim-level generator metrics (Completeness, Conciseness, and Verifiableness). On a human-annotated benchmark spanning eight datasets, \algname{} achieves higher correlation with human judgments than four existing RAG evaluation metrics, including RAGEval and RAGChecker, proving its effectiveness as a reliable, automated evaluation framework. Code and data are publicly available at \url{https://github.com/DISL-Lab/Q-CaRE-COLM-26}.

\vspace{4mm}

\noindent
\begin{minipage}[t]{0.7\textwidth}
{\small
\textbf{Date:} July 31, 2026 \par
\textbf{Correspondence:} Hwanjun Song at {\color{metablue}\href{mailto:songhwanjun@kaist.ac.kr}{songhwanjun@kaist.ac.kr}} \par
\textbf{First Author:} Jeonghwan Choi at {\color{metablue}\href{mailto:hwani.choi@kaist.ac.kr}{hwani.choi@kaist.ac.kr}} \par
\textbf{Code \& Data:} {\color{metablue} \url{https://github.com/DISL-Lab/Q-CaRE-COLM-26}}
}
\end{minipage}
\hfill
\begin{minipage}[t]{0.27\textwidth}
\vspace*{-0.3cm}
\raggedleft
\includegraphics[width=1.0\linewidth]{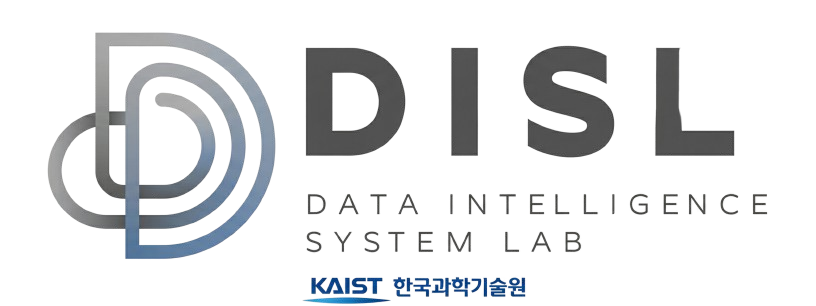}
\end{minipage}

\end{tcolorbox}
}

\begin{document}

\lhead{Published as a conference paper at COLM 2026}
\thispagestyle{empty}

\vspace*{-1.3cm}
\customabstractpage

\section{Introduction}
\label{sec:introduction}

Retrieval-augmented generation (RAG) improves the factuality of large language models (LLMs) by grounding responses in retrieved evidence \citep{gao2023retrieval, salemi2024retqual}. However, evaluating RAG systems remains challenging because failures can arise from retrieval, generation, or their interaction \citep{jon2024ares}. While prior work has proposed end-to-end evaluation frameworks for the entire pipeline \citep{es2024ragas, Ru2024ragchecker}, real-world evaluation is further complicated by the diversity of user queries, ranging from {"close-ended"} fact-seeking questions to {"open-ended"} requests requiring multi-faceted explanations. To be reliable in practice, an evaluation framework must therefore provide consistent measurements across this spectrum, a property we term \emph{query-agnostic} evaluation.

\begin{wraptable}{r}{0.5\textwidth}  
\centering
\scriptsize
\setlength{\tabcolsep}{1.2pt}
\vspace*{0.1cm}                       
\begin{tabular}{lcccc}
\toprule
{Method} & {Evaluation} & {Required} & {Evaluation} & {Query} \\
 & Target & {Reference} & {Granularity} & {Coverage} \\
\midrule
ARES       & \cellcolor{lightgreen}Ret. \& Gen. & \cellcolor{lightgreen}Free   & \cellcolor{lightred}Coarse  & \cellcolor{lightred}Close \\
RAGAs      & \cellcolor{lightgreen}Ret. \& Gen. & \cellcolor{lightgreen}Free   & \cellcolor{lightred}Coarse  & \cellcolor{lightred}Close \\
RAGEval    & \cellcolor{lightgreen}Ret. \& Gen. & \cellcolor{lightred}Chunk   & \cellcolor{lightgreen}Fine   & \cellcolor{lightred}Close \\
RAGChecker & \cellcolor{lightgreen}Ret. \& Gen. & \cellcolor{lightred}Answer  & \cellcolor{lightgreen}Fine   & \cellcolor{lightred}Close \\
\midrule
DoRAG      & \cellcolor{lightgreen}Ret. \& Gen. & \cellcolor{lightgreen}Free   & \cellcolor{lightred}Coarse  & \cellcolor{lightred}Open \\
\midrule
\algname{}(Ours) & \cellcolor{lightgreen}Ret. \& Gen. & \cellcolor{lightgreen}Free & \cellcolor{lightgreen}Fine & \cellcolor{lightgreen}Close \& Open \\
\bottomrule
\end{tabular}
\vspace*{-0.2cm}
\caption{Comparison of RAG evaluation methods along evaluation target, reference requirement, granularity, and query coverage (Ret.: Retrieval, Gen.: Generation).}
\label{tab:comparison}
\vspace*{-0.1cm}
\end{wraptable}

Achieving query-agnostic evaluation remains challenging because existing methods \citep{jon2024ares, es2024ragas, zhu2024rageval, Ru2024ragchecker, xie2024dorag} primarily evolve along two axes: \texttt{Reference-Dependence} (3rd column) and \texttt{Evaluation Granularity} (4th column), as summarized in Table \ref{tab:comparison}.
This leads to a structural \emph{trade-off} where metrics are specialized for different paradigms. Methods such as RAGEval \citep{zhu2024rageval} and RAGChecker \citep{Ru2024ragchecker} provide fine-grained diagnostics but rely on static gold answers or gold chunks, limiting them to close-ended queries. In contrast, reference-free approaches like DoRAG \citep{xie2024dorag} can handle open-ended queries but sacrifice granularity, producing signals much coarser than claim-level verification. Consequently, no existing method achieves fine-grained, reference-free evaluation while jointly supporting both close- and open-ended query types.

\begin{figure}[t]
\centering
\includegraphics[width=14cm]{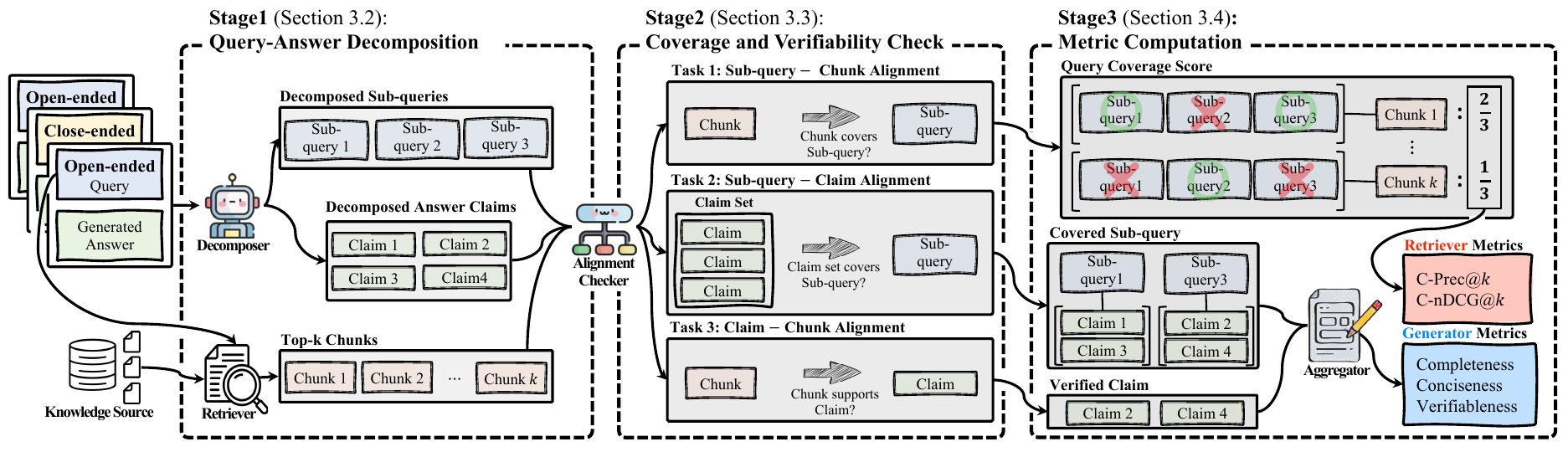}
\vspace*{-0.6cm}
\caption{Overview of \algname{}. It jointly evaluates retrieval and generation for both close- and open-ended queries by (Stage 1) decomposing queries and answers, (Stage 2) aligning them with retrieved chunks to assess coverage and verifiability, and (Stage 3) aggregating these signals into query-level metrics for retrieval and generation.}
\label{fig:overview}
\vspace*{-0.6cm}
\end{figure}

We contend that this bottleneck stems from a narrow view of evaluation that treats correctness as isolated checks, such as whether a chunk is relevant or a claim matches a reference, rather than asking ``{whether the query's original intent is satisfied}," which remains the same regardless of query type. 
To bridge this gap, we propose \algname{} (\underline{Q}uery \underline{C}overage and cl\underline{A}im ve\underline{R}ifiability for RAG \underline{E}valuation) in Figure \ref{fig:overview}, a framework that introduces a fundamental conceptual re-interpretation of answer correctness. Unlike existing methods, \algname{} establishes a unifying principle applicable to \emph{any} query type: a response is considered {"correct"} if it faithfully fulfills two criteria: \emph{(i) coverage}, meaning the answer addresses all information needs implied by the query, and \emph{(ii) verifiability}, meaning every statement in the answer is supported by the retrieved evidence. 

The query-agnosticism of \algname{} arises from its ability to translate any inquiry into these two universal signals---coverage and verifiability---through adaptive decomposition that adjusts to query complexity.
Specifically, we decompose queries into sub-queries and answers into atomic claims, evaluating coverage at the sub-query level and verifiability at the claim level using only retrieved evidence. This unified procedure applies equally to both close-ended and open-ended queries, thereby enabling a fully reference-free, fine-grained, and query-agnostic evaluation of RAG outputs.

Motivated by this intuition, we instantiate evaluation as a unified decomposition-based framework that operates in three stages:

\noindent{\textbf{Stage 1. Query-Answer Decomposition:}} It decomposes queries to sub-queries based on query's complexity, while answers into atomic answer claims for fine-grained evaluation.

\noindent{\textbf{Stage 2. Coverage and Verifiability Check:}} It performs three alignment check tasks over retrieved documents, decomposed sub-queries, and answer claims, enabling fully reference-free evaluation by verifying retrieval coverage of sub-queries, answer coverage over sub-queries, and claim-level answer verifiability against retrieved evidence.

\noindent{\textbf{Stage 3. Metric Computation}}: It combines atomic-level coverage and verifiability signals from Stage 2 to construct query-level evaluation dimensions that capture retrieval sufficiency and generation quality.

Through its three stages, \algname{} composes five complementary evaluation dimensions. On the retrieval side, it extends Precision@k and nDCG@k with coverage-aware soft relevance scores derived from sub-query coverage, replacing binary ground-truth relevance labels; we refer to these metrics as \emph{C-Prec@k} and \emph{C-nDCG@k}. On the generation side, it evaluates responses in terms of \emph{Completeness} (whether all sub-queries are addressed), \emph{Conciseness} (whether the answer avoids irrelevant claim), and \emph{Verifiableness} (whether answer claims are supported by retrieved evidence).

Our main contributions are summarized as: 

\noindent(1) We reinterpret answer correctness as the joint satisfaction of query coverage and  claim verifiability, providing a unified evaluation principle applicable to both close-ended and open-ended queries.

\noindent(2) We propose \algname{}, a decompositional framework that jointly evaluates retrieval and generation via query and answer decomposition, enabling end-to-end, fine-grained, and reference-free evaluation.

\noindent(3) We construct a human-annotated benchmark spanning diverse domains and query types for coverage- and verifiability-based evaluation.

\noindent(4) We demonstrate that \algname{} achieves the highest agreement with human judgments in RAG evaluation and that query coverage plays a critical role in reliable RAG benchmarking.
\vspace*{-0.4cm}
\section{Related Work}
\label{sec:realtedwork}
\vspace*{-0.05cm}

\noindent{\textbf{RAG Evaluation.}} Traditional methods rely on metrics such as Precision@k and nDCG@k for retrievers and EM and ROUGE for generators \citep{jarvelin2002cumulated, lin2004rouge, rajpurkar2016squad}. To enable semantic-aware evaluation, recent methods have adopted LLM-as-a-Judge approaches \citep{rau2024bergen, yu2024eval}, varying in supported query types, decomposition strategies, and reference requirements. ARES \citep{jon2024ares} and RAGAs \citep{es2024ragas} provide end-to-end pipeline evaluation but offer only coarse-grained assessment targeted at close-ended queries. RAGChecker \citep{Ru2024ragchecker} and RAGEval \citep{zhu2024rageval} introduce claim-level answer decomposition for fine-grained evaluation, yet depend on reference answers or human labeled chunks, and remain focused on close-ended. DoRAG \citep{xie2024dorag} targets open-ended queries through coverage-aware metrics for generators, but it only checks whether at least one retrieved chunk covers each sub-question, lacking fine-grained evaluation and favoring open-ended queries.

\smallskip
\noindent{\textbf{Decompose-based Evaluation.}} Evaluating complex QA is challenging because queries and answers often involve multiple aspects. Decompositional evaluation addresses this by breaking them into simpler components for finer assessment. On the query side, decomposition breaks complex queries into simpler sub-queries that can be independently addressed \citep{min2019multi, perez2020unsupervised}. Early work focused on multi-hop QA by decomposing queries into sequential reasoning steps \citep{wolfson2020break, khot2021text}. Recent methods generate targeted sub-queries to improve retrieval, where each sub-query captures a specific information need \citep{press2023selfask, shao2023iter}. On the answer-side, decomposition splits generated responses into atomic claims, enabling fine-grained factuality evaluation \citep{min2023factscore, wei2024longform}. This paradigm has been widely adopted for fact verification \citep{manakul2023selfcheckgpt}. Methods like FActScore \citep{min2023factscore} and SAFE \citep{wei2024longform} verify each claim against external evidence.
While effective, they typically apply decomposition in isolation, either on the retrieval side or on the generation side.

\vspace*{-0.05cm}
\section{Q-CARE Framework}
\label{sec:method}
\vspace*{-0.05cm}

We introduce \algname{} (in Figure \ref{fig:overview}) that quantifies query coverage and claim verifiability at the atomic level and aggregates them into retrieval and generation metrics.

\vspace*{-0.05cm}
\subsection{Target Evaluation Dimensions}
\label{subsec:dimensions}
\vspace*{-0.05cm}

We first outline the evaluation dimensions for RAG, emphasizing query coverage and claim-level verifiability beyond existing coarse metrics.
Let $\mathcal{Q}$ be a query. A RAG system retrieves top-$k$ chunks $\mathcal{C}=\{c_1,\ldots,c_k\}$ and generates an answer $\mathcal{A}=\text{Generator}(\mathcal{Q}\mid\mathcal{C})$. Through query–answer decomposition (Section \ref{subsec:decomposition}) and coverage and verifiability checks (Section \ref{subsec:alignment}), we derive retrieval metrics reflecting chunk-level query coverage and generation metrics reflecting answer-level query coverage and claim-level grounding. 

\noindent\textbf{Retriever Metrics.} We replace human relevance labels with coverage-aware relevance scores derived from how many sub-queries each chunk covers.
\begin{itemize}[leftmargin=*, noitemsep, topsep=0pt]
\item \emph{C-Prec@k}: The proportion of retrieved chunks containing query-relevant information, measured by sub-query coverage.
\item \emph{C-nDCG@k}: The ranking quality of retrieved chunks based on sub-query coverage, reflecting how well higher-ranked chunks cover the query.
\end{itemize}

\noindent\textbf{Generator Metrics.} Using atomic-level coverage and verifiability signals, we define three evaluation dimensions for generated answers.
\begin{itemize}[leftmargin=*, topsep=0pt]
\item \emph{Completeness}: The proportion of sub-queries addressed by the generated answer.
\item \emph{Conciseness}: The proportion of answer claims that are relevant to the covered sub-queries.
\item \emph{Verifiableness}: The proportion of answer claims that can be verified by the retrieved evidence.
\end{itemize}

\subsection{Query-Answer Decomposition}
\label{subsec:decomposition}
The first stage of \algname{} performs query–answer decomposition, breaking queries and generated answers into minimal, self-contained units without information loss (see Table \ref{tab:decomposition_example} in the Appendix for a detailed example).
Given an input text $\mathcal{X}$ (either query $\mathcal{Q}$ or answer $\mathcal{A}$), Decomposer produces a set of atomic units:
\begin{equation}
\text{Decomposer}(\mathcal{X}) \rightarrow \{x_1, x_2, ..., x_l\}.
\label{eq:decomp}
\end{equation}
To ensure the decomposed set is complete yet non-redundant, we enforce two constraints:
\begin{itemize}[leftmargin=*, noitemsep, topsep=0pt]
\item \textit{Exclusiveness}: Units must be mutually exclusive (\emph{i.e.}, $x_i \cap x_j = \emptyset$ for all $i \neq j$).
\item \textit{Completeness}: Units must collectively cover all information in the input (\emph{i.e.}, $\bigcup_{i=1}^{l} x_i = \mathcal{X}$).
\end{itemize}
While using an LLM-based prompting strategy,\footnote{As a fully automated metric, we use a single, replaceable LLM for all stages, using Qwen3-30B-A3B-Inst by default, and results with additional backbones in Section \ref{sec:baseline}.} the prompt constraints differ by target.

\noindent\textbf{Query Decomposition.} We apply Eq.~\eqref{eq:decomp} to obtain a set of sub-queries $\mathcal{Q} = \{q_1, q_2, ..., q_m\}$, where each $q_i \in \mathcal{Q}$ captures a distinct aspect of the original query. To handle both query types, we prompt the LLM to first identify the query type and then apply the corresponding strategy (refer to the detailed prompt in Appendix \ref{sec:prompts}). For \textit{single-focus} queries, which are close-ended, we keep them as-is. For \textit{multi-aspect} queries, which are open-ended, we decompose them into minimal perspectives. For \textit{multi-step reasoning} queries (\emph{e.g.}, multi-hop), which span both query types and require information from multiple chunks, we decompose them into sequential logical steps, enabling fine-grained evaluation and covering both query types. The effectiveness of decomposition is evaluated via an ablation study in Section \ref{benchmarking}.

\noindent\textbf{Answer Decomposition.} Given the generated answer $\mathcal{A}$, we apply Eq.~\eqref{eq:decomp} to obtain a set of answer claims $\mathcal{A} = \{a_1, a_2, ..., a_n\}$, where each $a_j \in \mathcal{A}$ represents a single, verifiable piece of information (refer to the detailed prompt in Appendix \ref{sec:prompts}). Each answer claim must be self-contained and understandable without the original query, as incomplete answer claims hinder accurate alignment checks. For example, given the query ``who plays Harley on Stuck in the Middle?'' and the answer ``Jenna Ortega,'' we extract the answer claim ``Jenna Ortega plays Harley on Stuck in the Middle'' from the answer rather than keeping it as-is.

\subsection{Coverage and Verifiability Check}
\label{subsec:alignment}

The second stage of \algname{} performs atomic-level coverage and verifiability checks to assess query coverage and claim grounding using the decomposed units. The decomposed sub-queries and answer claims, together with the retrieved chunks, undergo an alignment check process, which determines their pairwise alignments.

While each alignment task serves a distinct purpose, all use a shared alignment check by the Checker LLM. Given any two sets $\mathcal{Y}$ and $\mathcal{Z}$ from $\{\mathcal{Q}, \mathcal{A}, \mathcal{C}\}$, the Checker examines all pairs in $\mathcal{Y} \times \mathcal{Z}$ and returns the set of aligned pairs:
\begin{equation}
\text{Checker}(\mathcal{Y}, \mathcal{Z}) \rightarrow E(\mathcal{Y}, \mathcal{Z}) \subseteq \mathcal{Y} \times \mathcal{Z},
\label{eq:checker}
\end{equation}
where $(y, z) \in E(\mathcal{Y}, \mathcal{Z})$ indicates that $y$ and $z$ are aligned. All assessments are performed via LLM prompts detailed in Appendix~\ref{sec:prompts}.

\noindent\textbf{Task 1: Sub-query -- Chunk Alignment.} This task identifies which retrieved chunks support which sub-queries. The $\text{Checker}(\mathcal{Q}, \mathcal{C})$ returns the set of aligned sub-query--chunk pairs:
\begin{equation}
E(\mathcal{Q}, \mathcal{C}) = \{(q, c) \mid q \in \mathcal{Q}, c \in \mathcal{C}, c \rightarrow q\},~~{\rm where}
\label{eq:query-chunk_align}
\end{equation}
$c \rightarrow q$ indicates that chunk $c$ contains sufficient information to answer the sub-query $q$.

\noindent\textbf{Task 2: Sub-query -- Claim Alignment.} This task determines which answer claims are relevant to addressing each sub-query. The $\text{Checker}(\mathcal{Q}, \mathcal{A})$ returns the set of aligned sub-query--claim pairs:
\begin{equation}
E(\mathcal{Q}, \mathcal{A}) = \{(q, a) \mid q \in \mathcal{Q}, a \in \mathcal{A}, a \rightarrow q\},~~{\rm where}
\label{eq:query-claim_align}
\end{equation}
$a \rightarrow q$ indicates that answer claim $a$ is relevant to answering the sub-query $q$. 

\noindent\textbf{Task 3: Claim -- Chunk Alignment.} This task verifies whether each answer claim is supported by the retrieved chunks, assessing the verifiability of each answer claim. The $\text{Checker}(\mathcal{A}, \mathcal{C})$ returns the set of aligned claim--chunk pairs:
\begin{equation}
E(\mathcal{A}, \mathcal{C}) = \{(a, c) \mid a \in \mathcal{A}, c \in \mathcal{C}, c \rightarrow a\},~~{\rm where}
\label{eq:claim-chunk_align}
\end{equation}
$c \rightarrow a$ indicates that chunk $c$ provides evidence supporting the answer claim $a$. 

\subsection{Metrics Computation}
\label{subsec:metric}
Lastly, \algname{} aggregates the atomic-level results in Eqs~\eqref{eq:query-chunk_align}--\eqref{eq:claim-chunk_align} and computes the five metrics, measuring query coverage for retrieval and completeness, conciseness, and verifiability for generation, enabling consistent evaluation across all query types.

\subsubsection{Coverage-aware Retriever Metrics}
\label{subsubsec:retriever}

Traditional metrics such as Precision@k and nDCG@k rely on human relevance labels and binary assumptions. While suitable for close-ended queries, these assumptions break down in realistic RAG settings, where large-scale high-quality annotation is infeasible \citep{rassin2024evaluating, Takehi2025} and binary labels cannot capture the degree of query coverage in each chunk. We thus modify the canonical metrics to reflect chunk-level query coverage.

For each chunk $c$, we compute its \emph{query coverage score} using aligned sub-query--chunk pairs:
\begin{equation}
r = \frac{|\{q  \mid (q, c) \in E(\mathcal{Q}, \mathcal{C})\}|}{|\mathcal{Q}|} \in [0, 1], ~~{\rm where}
\label{eq:soft_relevance_label}
\end{equation}
$r$ denotes the fraction of sub-queries in $\mathcal{Q}$ that are covered by the chunk $c$. Using these soft coverage labels, we modify Precision and nDCG to be fully reference-free by replacing human relevance labels with atomic-level automatic alignment.

\noindent\textbf{C-Prec@k.} C-Prec@k (k: \# of retrieved chunks) replaces binary relevance in Precision@k with soft coverage scores $r_i$, enabling proportional credit for partially relevant chunks:
\begin{equation}
\text{C-Prec@k} = \frac{\sum_{i=1}^{k} r_i}{k}.
\label{eq:c_pre}
\end{equation}
\noindent\textbf{C-nDCG@k.} While C-Prec@k measures the average query coverage within the top-$k$ retrieved chunks, C-nDCG@k additionally accounts for their ranking order by weighting soft coverage scores $r_i$ according to their retrieval positions:
\begin{equation}
\text{C-nDCG@k} = \frac{\text{DCG@k}}{\text{IDCG@k}}, ~~
{\rm where}~~ \text{DCG@k} = \sum_{i=1}^{k} \frac{r_i}{\log_2(i+1)}
\label{eq:c_ndcg}
\end{equation}
and IDCG@k is the ideal DCG obtained by sorting chunks by their query coverage scores.

\subsubsection{Multi-aspect Generation Metrics}
\label{subsubsec:generator}
We move beyond surface-level correctness and assess generation quality in terms of coverage and verifiability. Specifically, we decompose generation evaluation into three complementary dimensions: {Completeness}, {Conciseness}, and {Verifiableness}.

\noindent\textbf{Completeness and Conciseness.} Completeness measures whether all sub-queries are addressed, while conciseness penalizes unnecessary claims. Since higher verbosity can increase completeness by covering more sub-queries \citep{song2024finesure}, these dimensions must be evaluated jointly.
To compute the final answer-level metrics, we perform an LLM-based coverage check to verify whether the answer claims aligned with each sub-query in Eq.~\eqref{eq:query-claim_align} are sufficient to answer that sub-query, as relevance alone does not ensure that the sub-query is fully addressed. Accordingly, for each sub-query $q$, we collect the set of relevant answer claims, $R(q) = \{a \mid (q, a) \in E(\mathcal{Q}, \mathcal{A})\}$. We then prompt the same LLM to determine whether the claims in $R(q)$ are sufficient to answer $q$ (see the prompt in Table \ref{tab:prompt_sf_coverage}). As a result, we obtain the set of fully covered sub-queries $\mathcal{Q}_{\mathrm{cov}}(\subseteq \mathcal{Q}$), where $q \in \mathcal{Q}_{\mathrm{cov}}$ denotes a sub-query that can be fully answered by its associated claims $R(q)$.

Therefore, completeness is defined as the fraction of sub-queries that can be fully answered by their associated answer claims:
\begin{equation}
\text{Completeness} = {|\mathcal{Q}_{\text{cov}}|}/{|\mathcal{Q}|}.
\label{eq:completeness}
\end{equation}
On the other hand, conciseness is defined as the fraction of answer claims that contribute to fully covered sub-queries:
\begin{equation}
\text{Conciseness} = {|\mathcal{A}_{\mathrm{cov}}|}/{|\mathcal{A}|}, ~~\text{where}~~ \mathcal{A}_{\mathrm{cov}}=
\{ a \mid (q,a) \in E(\mathcal{Q}, \mathcal{A}) \text{ s.t. } \exists q \in\mathcal{Q}_{\mathrm{cov}}\}.
\label{eq:conciseness}
\end{equation}
Here, $E(\mathcal{Q}, \mathcal{A})$ denotes the set of aligned sub-query–claim pairs in Task 2.

\noindent\textbf{Verifiableness.} Completeness alone does not guarantee correctness, as answer claims must also be supported by evidence. Verifiableness thus evaluates whether answer claims can be verified against the retrieved chunks, penalizing unsupported or hallucinated claims. It is defined as the fraction of answer claims that are grounded in the retrieved chunks:
\begin{equation}
\text{Verifiableness} = \frac{|\{a \mid (a, c) \in E(\mathcal{A}, \mathcal{C})\}|}{|\mathcal{A}|}, ~~{\rm where}
\label{eq:verifiableness}
\end{equation}
$E(\mathcal{A}, \mathcal{C})$ is the set of aligned answer claim--chunk pairs in Task 3. That is, the generation metrics are complementary, jointly capturing coverage, conciseness, and factual grounding of the generated answers in a unified manner, without relying on reference answers.
\section{Benchmark Construction} 
\label{subsec:benchmark}

Our framework evaluates RAG systems based on coverage and verifiability, which requires human annotations on sub-query coverage and answer-claim grounding. Since no existing benchmark provides these annotations, we construct a new benchmark spanning diverse domains and both close-ended and open-ended query types.


\noindent\textbf{Source Dataset.} We select eight datasets across general, news, finance, biomedical, science, and technology domains, categorized by query complexity to assess query type-agnostic capability:
\textit{Single-focus} (NQ, NewsQA),  \textit{Multi-hop reasoning} (HotpotQA, FinQA), and \textit{Multi-aspect} (PubMedQA, LoTTE-Science, LoTTE-Technology, ELI5).
Single-focus and multi-hop queries typically have definitive answers, whereas multi-aspect queries admit multiple valid responses. Accordingly, we construct close-ended data from NQ, NewsQA, HotpotQA, and FinQA, and open-ended data from PubMedQA, LoTTE-Science, LoTTE-Technology, and ELI5. Queries are filtered using an LLM to retain only those matching the intended query type, with prompts provided in Appendix \ref{sec:prompts}.

\noindent\textbf{Query-Answer Sampling.} We sample 160 queries from each query type (\emph{i.e.}, close-ended and open-ended), sampling an equal number of queries per dataset, for human correlation analysis. For each query, we retrieve top-10 chunks using BM25 from the corresponding corpus and generate responses using 8 LLMs of varying sizes including open-source and proprietary models: Gemma-3-4B-IT, Gemma-3-27B-IT, Qwen3-4B-Inst., Qwen3-30B-A3B-Inst., GPT-oss-20B, GPT-5, Claude-Sonnet-4, and Gemini-2.5-pro (see Appendix~\ref{sec:llm_config.} for details).


\begin{wraptable}{r}{0.54\textwidth}
\centering
\scriptsize
\setlength{\tabcolsep}{0.1pt}
\begin{tabular}{llcc}
\toprule
Stage & Task & Close-ended & Open-ended \\
\midrule
\multirow{2}{*}{Decomposition}
& Query  & 99.3\% & 100\%  \\
& Answer & 98.9\% & 97.9\% \\
\midrule
\multirow{3}{*}{\shortstack[l]{Alignment\\Check}}
& Task 1 (Sub-query -- Chunk) & 0.837 & 0.801 \\
& Task 2 (Sub-query -- Claim) & 0.891 & 0.883 \\
& Task 3 (Claim -- Chunk)     & 0.873 & 0.890 \\
\bottomrule
\end{tabular}
\vspace{-0.25cm}
\caption{Human annotation quality measured by rate of three annotators agreeing on decomposition results and Fleiss' kappa.}
\label{tab:agreement}
\vspace*{-0.2cm}
\end{wraptable}
This high IAA confirms reliable human annotation at the task level. Using these annotations, we compare how well different evaluation methods correlate with human judgments. More details on human annotation are provided in Appendix~\ref{sec:human_annotation}.

\noindent\textbf{Human Annotation.} We collect human annotations on the constructed dataset via Amazon Mechanical Turk (MTurk) under the \algname{}'s pipeline, instead of LLM judges.\footnotemark[2] This results in 567 sub-queries and 7,907 answer claims across the benchmark (see Appendix~\ref{sec:benchmark} for details). For alignment check tasks (Section~\ref{subsec:alignment}), three annotators independently label each instance for Tasks 1--3. Table \ref{tab:agreement} shows the inter-annotator agreement (IAA) among human annotators for the three tasks, achieving Fleiss' kappa scores ranging from 0.801 to 0.891.

\footnotetext[2]{While most tasks rely on humans, we use GPT-5 only for query–answer decomposition (Stage 1), since prior work shows it can be reliably handled by strong LLMs \citep{min2023factscore, Ru2024ragchecker}. This is also confirmed by our verification, with over 97\% agreement with human judgments in Table \ref{tab:agreement}.} 
\footnotetext[3]{For brevity, C-Prec and C-nDCG are denoted as Prec. and nDCG., respectively, in Tables \ref{tab:main_correlation} and \ref{tab:backbone}.}

\section{Evaluation}
\label{sec:evaluation}

\begin{wraptable}{r}{0.65\textwidth}
\centering
\scriptsize
\setlength{\tabcolsep}{0.9pt}
\vspace*{-0.3cm}
\begin{tabular}{lccccccccccc}
\toprule
\multirow{3}{*}{Method} & \multicolumn{5}{c}{Close-ended} & \multicolumn{5}{c}{Open-ended} \\
\cmidrule(lr){2-6} \cmidrule(lr){7-11}
 & \multicolumn{2}{c}{Retriever} & \multicolumn{3}{c}{Generator} & \multicolumn{2}{c}{Retriever} & \multicolumn{3}{c}{Generator} \\
\cmidrule(lr){2-3} \cmidrule(lr){4-6} \cmidrule(lr){7-8} \cmidrule(lr){9-11}
 & Prec. & nDCG. & Comp. & Conc. & Veri. & Prec. & nDCG. & Comp. & Conc. & Veri. \\
\midrule
RAGEval   & 0.29\sig & N/A      & 0.27\sig & N/A      & 0.22\sig & 0.35\sig & N/A      & 0.13\sig & N/A      & 0.05\sig \\
RAGAs     & 0.24\sig & N/A      & 0.44\sig & N/A      & 0.31\sig & 0.22\sig & N/A      & 0.28\sig & N/A      & 0.14\sig \\
RAGCheck. & 0.32\sig & 0.32\sig & 0.42\sig & 0.33\sig & 0.28\sig & 0.22\sig & 0.19\sig & 0.17\sig & 0.20\sig & 0.56\sig \\
DoRAG     & N/A      & N/A      & 0.28\sig & N/A      & N/A      & N/A      & N/A      & 0.23\sig & N/A      & N/A      \\
\midrule
\textbf{\algname{}} & \textbf{0.55} & \textbf{0.43} & \textbf{0.52} & \textbf{0.47} & \textbf{0.40} & \textbf{0.49} & \textbf{0.43} & \textbf{0.47} & \textbf{0.36} & \textbf{0.69} \\
\bottomrule
\end{tabular}
\vspace{-0.25cm}
\caption{Agreement between human judgments and five RAG evaluation metrics. N/A denotes metrics not computable due to method-specific design assumptions. $^*$: $p<0.01$, paired bootstrap test against all baselines.}
\label{tab:main_correlation}
\vspace*{-0.3cm}
\end{wraptable}

We compare \algname{} over four RAG metrics (Section \ref{sec:baseline}), including RAGEval, RAGAs, RAGCheck, and DoRAG, as well as conventional metrics (Section \ref{sec:impact_coverage_aware}), including Precision and nDCG for retrieval, and ROUGE-L, EM, and ACC for generation. All LLM-based methods use Qwen3-30B-A3B-Inst. as the backbone for consistency. Details can be found in {Appendix \ref{sec:detail_conventional}}. We evaluate them on our benchmark datasets described in Section~\ref{subsec:benchmark} and primarily report agreement with human judgments using the \emph{Pearson correlation coefficient}, following the existing work \citep{song2024finesure, Ru2024ragchecker}. Higher values indicate stronger agreement with human judgments.

\subsection{Main Results}
\label{sec:baseline}

\noindent\textbf{Comparison with LLM-based Metrics.} 
Table \ref{tab:main_correlation} reports the agreement between dimension-wise human judgments and five RAG evaluation metrics, including \algname{}, across two retrieval metrics, C-Prec@k and C-nDCG@k\footnotemark[3], and three generation metrics, Completeness (Comp.), Conciseness (Conc.), and Verifiableness (Veri.). Overall, \algname{} achieves the highest correlation with human judgments across all dimensions and across both query types, showing \algname{}'s query-agnostic evaluation capability. In contrast, other methods show notable performance degradation from close-ended to open-ended queries. For instance, RAGAs and RAGChecker achieve moderate correlation on Completeness for close-ended queries, but their performance drops on open-ended queries. We further analyze latency in Appendix \ref{sec:latency}, showing that \algname{} does not add significant runtime overhead over baselines.




\noindent\textbf{Impact of Backbone LLMs.} We analyze how the choice of backbone models impacts the performance of \algname{} according to their size and family. 
\begin{wraptable}{r}{0.6\textwidth}
\centering
\scriptsize
\setlength{\tabcolsep}{0.9pt}
\vspace*{-0.4cm}
\begin{tabular}{lccccccccccc}
\toprule
\multirow{3}{*}{Model} & \multicolumn{5}{c}{Close-ended} & \multicolumn{5}{c}{Open-ended} \\
\cmidrule(lr){2-6} \cmidrule(lr){7-11}
 & \multicolumn{2}{c}{Retriever} & \multicolumn{3}{c}{Generator} & \multicolumn{2}{c}{Retriever} & \multicolumn{3}{c}{Generator} \\
\cmidrule(lr){2-3} \cmidrule(lr){4-6} \cmidrule(lr){7-8} \cmidrule(lr){9-11}
 & Prec. & nDCG. & Comp. & Conc. & Veri. & Prec. & nDCG. & Comp. & Conc. & Veri. \\
\midrule
GPT5-mini    & 0.56 & 0.54 & 0.51 & 0.50 & 0.45 & 0.42 & 0.42 & 0.23 & 0.36 & 0.71 \\
Llama3.1-8B  & 0.55 & 0.40 & 0.14 & 0.16 & 0.07 & 0.47 & 0.41 & 0.11 & 0.08 & 0.15 \\
Qwen3-30B    & 0.55 & 0.43 & 0.52 & 0.47 & 0.40 & 0.49 & 0.48 & 0.47 & 0.36 & 0.69 \\
Llama3.3-70B & 0.63 & 0.50 & 0.44 & 0.37 & 0.50 & 0.57 & 0.50  & 0.31 & 0.30 & 0.65 \\
\bottomrule
\end{tabular}
\vspace{-0.25cm}
\caption{Agreement between human judgements and \algname{}'s variants using different LLMs.}
\label{tab:backbone}
\vspace*{-0.25cm}
\end{wraptable}
Table \ref{tab:backbone} reports the agreement with human judgments when varying the LLMs used in \algname{}. Smaller models such as Llama-3.1-8B-Inst. exhibit weak correlation, particularly for open-ended queries, whereas larger models achieve substantially higher correlation across both query types. This indicates that \algname{} benefits from stronger backbone models. 
{Among larger models, Qwen3-30B-A3B-Inst. offers a practical balance between performance and computational cost, achieving comparable performance to Llama3.3-70B. despite using substantially fewer parameters.}

\subsection{Comparison with Traditional Metrics}
\label{sec:impact_coverage_aware}

Beyond recent RAG metrics, we examine whether traditional retrieval and generation metrics capture coverage-aware human judgments. While human evaluation considers whether all query aspects are addressed, existing metrics largely rely on binary relevance or reference-based similarity. We thus compare traditional metrics with \algname{}. 
{Here, correlations are computed using a human composite score averaged over Completeness, Conciseness, and Verifiability (see Appendix \ref{sec:composite}), as traditional metrics output a single score.}

\begin{wraptable}{r}{0.5\textwidth}
\centering
\scriptsize
\setlength{\tabcolsep}{7.7pt}
\vspace*{-0.2cm}
\begin{tabular}{lcccc}
\toprule
\multirow{2}{*}{Method} & \multicolumn{2}{c}{Close-ended} & \multicolumn{2}{c}{Open-ended} \\
\cmidrule(lr){2-3} \cmidrule(lr){4-5}
 & Prec. & nDCG. & Prec. & nDCG. \\
\midrule
Default                 & 0.286          & 0.141          & 0.264          & 0.331          \\
\textbf{Coverage-aware} & \textbf{0.348} & \textbf{0.372} & \textbf{0.391} & \textbf{0.431} \\
\bottomrule
\end{tabular}
\vspace{-0.25cm}
\caption{Agreement between human composite scores and retrieval metrics, comparing traditional (``Default'') metrics and \algname{} (``Coverage-aware'').}
\label{tab:retriever_correlation}
\vspace*{-0.25cm}
\end{wraptable}

\noindent{\textbf{Retriever Metrics.}} Table \ref{tab:retriever_correlation} compares traditional retrieval metrics (Default) with \algname{} (Coverage-aware). Default uses canonical Precision@$k$ and nDCG@$k$ computed from human binary relevance labels, whereas Coverage-aware uses our proposed coverage-aware metrics without such labels (\emph{i.e.}, C-Prec@$k$ and C-nDCG@$k$). The results reveal that traditional metrics exhibit weak correlation with composite human scores, whereas the coverage-aware metrics achieve substantially higher agreement across both query types. This indicates that accounting for query coverage is essential for aligning retrieval evaluation with human judgments.

\noindent{\textbf{Generation Metrics.}} Table \ref{tab:generator} compares the traditional generation metric (EM, Acc, ROUGE-L) and \algname{}. The three traditional metrics are reference-based and thus rely on the original reference answers provided in the source datasets. In contrast, \algname{} is fully reference-free, and we compare it using a composite score obtained by averaging its three generation dimensions. The results show that traditional metrics achieve reasonable correlation on close-ended queries, but fail to generalize to open-ended queries due to their reliance on incomplete reference answers. 
\begin{wraptable}{r}{0.5\textwidth}
\centering
\scriptsize
\setlength{\tabcolsep}{2.5pt}
\vspace*{-0.15cm}
\begin{tabular}{l*{4}{>{\centering\arraybackslash}p{1.22cm}}}
\toprule
 & \!\!Exact Match\!\! & Accuracy & \!\!\!ROUGE-L\!\!\! & \textbf{\algname{}} \\
\midrule
Close-ended &  0.244 & 0.369 & 0.357 & \textbf{0.547} \\
Open-ended  & -0.062 & 0.014 & 0.209 & \textbf{0.407} \\
\bottomrule
\end{tabular}
\vspace{-0.25cm}
\caption{Agreement between human composite scores and generation metrics, comparing traditional (EM, Acc, ROUGE-L) and \algname{} (its composite score).}
\label{tab:generator}
\vspace*{-1.5cm}
\end{wraptable}
Notably, \algname{} outperforms reference-based metrics even on close-ended queries, highlighting the importance of coverage awareness beyond open-ended settings.

\section{Benchmarking with \algname{}}
\label{benchmarking}
\vspace{-0.1cm}

We apply \algname{} to benchmark existing RAG systems, enabling stage-wise evaluation of retrieval and generation across 21 retrievers and 8 generators on our dataset. The benchmarking is conducted on 400 queries per query type, randomly sampled from the source datasets described in Section \ref{subsec:benchmark}. Details on retrieval and generator selection, as well as query sampling, are in Appendix \ref{sec:benchmark_config}.



\begin{figure*}[!t]
\centering
\includegraphics[width=14cm]{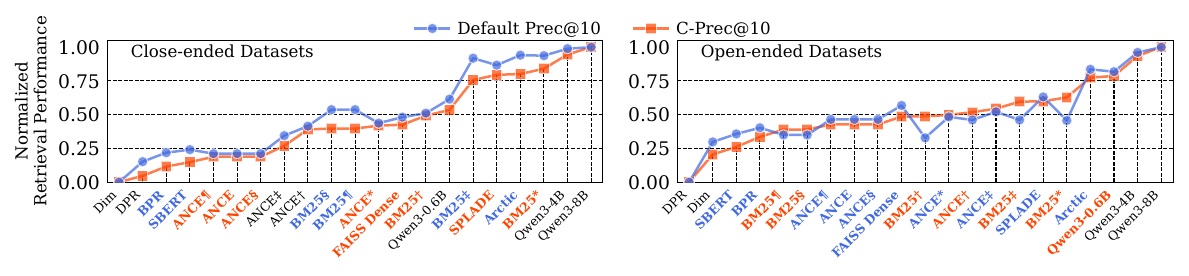}
\vspace*{-1cm}
\caption{Retriever performance with Default Precision@10 and Coverage-aware Precision@10, sorted by the latter. Retriever names are colored {\color[HTML]{4169E1}blue} if ranked higher by Default Precision@10, {\color[HTML]{FF4500}orange} if ranked higher by Coverage-aware Precision@10. Superscripts denote query expansion (\textsuperscript{\textdagger}HyDE, \textsuperscript{\textdaggerdbl}Q2D, \textsuperscript{*}MuGI) and reranking (\textsuperscript{\textsection}monoT5, \textsuperscript{\textparagraph}ELECTRA). Results with Default and Coverage-aware nDCG@$k$ is in Appendix \ref{sec:benchmark_config}.}
\label{fig:retriever_precision}
\vspace*{-0.3cm}
\end{figure*}

\noindent{\textbf{Retriever Benchmarking.}} Figure \ref{fig:retriever_precision} compares 21 retrieval strategies, including canonical sparse and dense methods and their combinations with query expansion and reranking. To facilitate clearer ranking comparisons, the retrieval performance is normalized across methods. We additionally report the default metric as a reference, enabling direct comparison with our coverage-aware metric and revealing rank discrepancies (in blue or orange).

First, we observe notable discrepancies between Default and Coverage-aware retrieval metrics, with rank changes for 14 and 17 retrievers on close-ended and open-ended datasets, respectively. In particular, sparse retrievers (\emph{e.g.}, BM25), as well as retrievers combined with query expansion, tend to be overestimated on close-ended queries but underestimated on open-ended queries under Default metrics.
Second, Coverage-aware metrics yield more consistent retriever rankings across close-ended and open-ended queries, increasing the Spearman rank correlation from $0.55$ to $0.89$. Together, these results indicate that system-level analysis can be misleading when query coverage is not explicitly accounted for.

\begin{figure*}[!t]
\centering
\includegraphics[width=14cm]{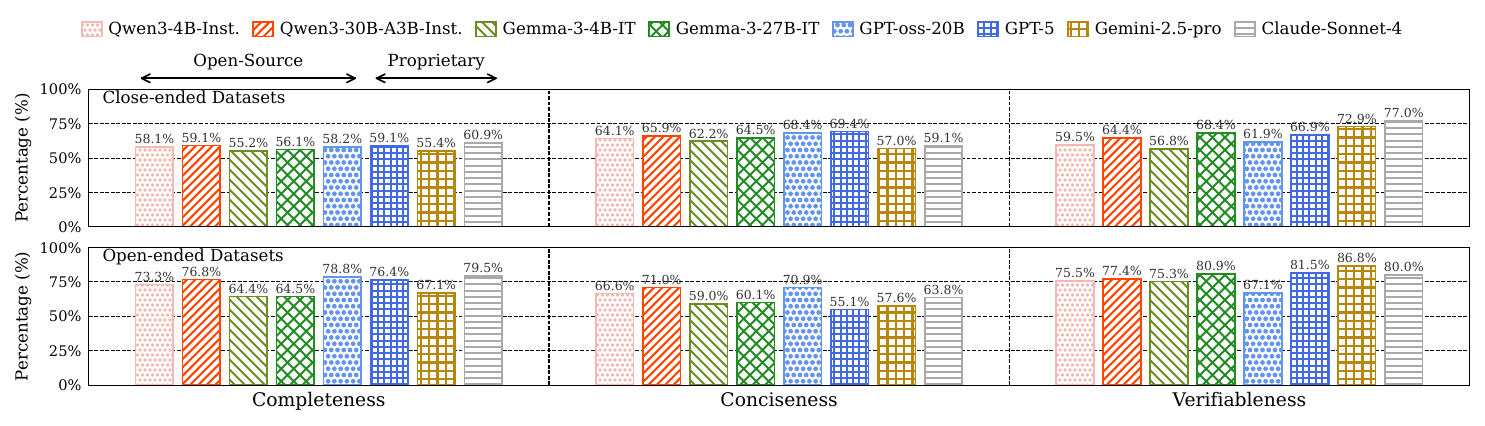}
\vspace*{-0.85cm}
\caption{Generator evaluation results using \algname{} across 8 LLMs on close-ended and open-ended datasets. We evaluate Completeness, Conciseness, and Verifiableness. See Appendix \ref{sec:benchmark_config} for the detailed LLM configuration.}
\label{fig:generator_eval}
\vspace*{-0.5cm}
\end{figure*}

\noindent{\textbf{Generator Benchmarking.}} Figure \ref{fig:generator_eval} compares the generation performance of eight RAG systems using different LLMs as generators, with the canonical BM25 retriever fixed to isolate retrieval effects.

\noindent$\bullet$~\emph{Completeness.} Proprietary LLMs, such as GPT-5 and Claude-Sonnet-4 achieve the highest completeness on both close-ended and open-ended queries, demonstrating robust coverage across query types. However, the open-source Gemma-family lags behind other models on open-ended datasets.

\noindent$\bullet$~\emph{Conciseness.} LLMs exhibit divergent trends between close-ended and open-ended queries. Qwen3-30B-A3B-Inst substantially improves conciseness on open-ended queries. In contrast, GPT-5 shows the largest drop in conciseness from close-ended to open-ended queries, likely reflecting a trade-off in favor of completeness in the open-ended setup.

\noindent$\bullet$~\emph{Verifiableness.} Claude-Sonnet-4 achieves the highest scores on close-ended queries, while Gemini-2.5-Pro performs best on open-ended ones. In contrast, {GPT-oss-20B} shows lower verifiableness on open-ended queries, suggesting greater reliance on internal knowledge.

These results indicate that no single LLM dominates across all evaluation dimensions. Instead, current LLMs show different performance dominance across evaluation dimensions, and the difference in the dominance patterns is also influenced by the query type whether it is open-ended or close-ended. This variability underscores the necessity of fine-grained, multi-dimensional evaluation.

\begin{table}[t]
\centering
\scriptsize
\setlength{\tabcolsep}{6.8pt}
\begin{tabular}{l cccc cccc}
\toprule
\multirow{2}{*}{Method} & \multicolumn{4}{c}{\textbf{Faithfulness}} & \multicolumn{4}{c}{\textbf{Completeness}} \\
\cmidrule(lr){2-5} \cmidrule(lr){6-9}
 & CLAPNQ & FIQA & GOVT & CLOUD & CLAPNQ & FIQA & GOVT & CLOUD \\
\midrule
RougeL   & 0.490$^{***}$ & 0.430$^{***}$ & 0.519$^{***}$ & 0.333$^{**}$  & 0.447$^{***}$ & 0.308$^{**}$  & 0.322$^{**}$  & 0.374$^{***}$ \\
BKPrec   & 0.681$^{***}$ & 0.491$^{***}$ & 0.698$^{***}$ & 0.542$^{***}$ & 0.611$^{***}$ & 0.322$^{**}$  & 0.403$^{***}$ & 0.457$^{***}$ \\
BRec     & 0.206         & 0.275$^{*}$   & 0.281$^{*}$   & 0.264$^{*}$   & 0.250$^{*}$   & 0.342$^{**}$  & 0.135         & 0.384$^{***}$ \\
RB\textsubscript{alg} & 0.567$^{***}$ & 0.422$^{***}$ & 0.380$^{**}$ & 0.271$^{*}$ & 0.529$^{***}$ & 0.306$^{**}$ & 0.262$^{*}$ & 0.405$^{***}$ \\
RL\textsubscript{F}   & 0.779$^{***}$ & 0.555$^{***}$ & 0.391$^{***}$ & 0.431$^{***}$ & 0.711$^{***}$ & 0.379$^{***}$ & 0.269$^{*}$ & 0.428$^{***}$ \\
RB\textsubscript{llm} & 0.327$^{**}$ & 0.074 & 0.081 & 0.139 & 0.352$^{**}$ & 0.250$^{*}$ & 0.196 & 0.166 \\
\midrule
\algname{} (Qwen3-30B) & 0.728$^{***}$ & 0.622$^{***}$ & 0.689$^{***}$ & 0.562$^{***}$ & 0.675$^{***}$ & 0.484$^{***}$ & 0.427$^{***}$ & 0.361$^{***}$ \\
\algname{} (Qwen3-80B) & 0.731$^{***}$ & 0.642$^{***}$ & 0.700$^{***}$ & 0.618$^{***}$ & 0.680$^{***}$ & 0.373$^{**}$ & 0.444$^{***}$ & 0.483$^{***}$ \\
\bottomrule
\end{tabular}
\vspace{-0.25cm}
\caption{Spearman correlation with human scores on conversational RAG benchmarks from MTRAG. Baseline metrics follow the definitions in MTRAG. \algname{} is evaluated with two backbone LLMs, Qwen3-30B-A3B-Inst.\ (Qwen3-30B) and Qwen3-Next-80B-A3B-Inst.\ (Qwen3-80B). $^{***}p<.001$, $^{**}p<.01$, $^{*}p<.05$ (two-tailed test).}
\vspace*{-0.5cm}
\label{tab:mtrag}
\end{table}

\textbf{Generalization to Conversational RAG.}
To assess whether \algname{} extends beyond single-turn settings, we evaluate it on MTRAG \citep{katsis2025mtrag}, a multi-turn conversational RAG benchmark spanning four datasets (CLAPNQ, FIQA, GOVT, and CLOUD). Following the original setup, we compare \algname{}, with Qwen3-30B-A3B-Inst.\ and Qwen3-Next-80B-A3B-Inst.\ backbones, against the six baseline metrics reported in MTRAG on Faithfulness and Completeness, the two dimensions that align with our framework, reporting Spearman correlations with human scores.
\begin{wraptable}{r}{0.58\textwidth}
\centering
\scriptsize
\setlength{\tabcolsep}{2.8pt}
\begin{tabular}{lcccccc}
\toprule
\multirow{2}{*}{Method} & \multicolumn{3}{c}{Qwen3-30B} & \multicolumn{3}{c}{Qwen3-80B} \\
\cmidrule(lr){2-4} \cmidrule(lr){5-7}
 & Comp. & Conc. & Prec. & Comp. & Conc. & Prec. \\
\midrule
\multicolumn{7}{l}{\textit{Close-ended}} \\
w.\ Decomp.  & 0.347 & 0.426 & 0.623 & 0.483 & 0.494 & 0.647 \\
w/o Decomp.  & 0.291 & 0.381 & 0.444 & 0.232 & 0.314 & 0.166 \\
$\Delta$     & $+$0.056\sig & $+$0.045\sig & $+$0.179\sig & $+$0.251\sig & $+$0.180\sig & $+$0.481\sig \\
\midrule
\multicolumn{7}{l}{\textit{Open-ended}} \\
w.\ Decomp.  & 0.487 & 0.338 & 0.416 & 0.816 & 0.538 & 0.636 \\
w/o Decomp.  & 0.391 & 0.301 & 0.583 & 0.343 & 0.493 & 0.511 \\
$\Delta$     & $+$0.096\sig & $+$0.038\sig & $-$0.167\sig & $+$0.473\sig & $+$0.045 & $+$0.125\sig \\
\bottomrule
\end{tabular}
\vspace{-0.25cm}
\caption{Ablation study on query decomposition, evaluated on HotpotQA (close-ended) and ELI5 (open-ended). \sig $p < .05$ (Wilcoxon signed-rank test). Comp.: Completeness, Conc.: Conciseness, Prec.: C-Prec@10.}
\label{tab:ablation}
\end{wraptable}
As shown in Table~\ref{tab:mtrag}, \algname{} achieves the highest correlation on three of the four datasets in both dimensions without any modification to the framework. This indicates that the coverage and verifiability principle generalizes naturally to conversational settings, a scenario that fine-grained single-turn frameworks such as RAGChecker, RAGEval, and DoRAG do not support. Moreover, scaling the backbone from 30B to 80B further improves correlation on six of the eight dataset--dimension combinations, suggesting that \algname{}'s reliability in multi-turn settings also benefits from stronger backbones.

\textbf{Ablation on Query Decomposition.}
To validate the contribution of decomposition, we compare \algname{} with a variant that removes query decomposition (\textit{w/o Decomp.}), using the original query directly for all alignment tasks. We evaluate on HotpotQA (close-ended) and ELI5 (open-ended) with Qwen3-30B-A3B-Inst.\ and Qwen3-Next-80B-A3B-Inst.\ backbones.
As shown in Table~\ref{tab:ablation}, query decomposition consistently and significantly improves Completeness across both query types and both backbones ($+$0.056 to $+$0.473, all $p < .05$). With the stronger Qwen3-80B backbone, decomposition brings significant gains across all three metrics for close-ended queries, and achieves a particularly large improvement on Completeness for open-ended queries ($+$0.473, $p < .05$). These results confirm that query decomposition is a critical component of \algname{}'s evaluation quality, not merely an auxiliary processing step.

\vspace*{-0.15cm}
\section{Conclusion} \label{sec:conclusion} \vspace*{-0.05cm}
\vspace*{-0.2cm}
We present \algname{}, a novel query-agnostic RAG evaluation framework based on query coverage and claim verifiability. \algname{} decomposes queries and answers into sub-queries and atomic claims, using alignment checks with retrieved chunks to enable query-agnostic evaluation across close- and open-ended settings. Experiments show that \algname{} achieves higher agreement with human judgments than existing RAG metrics.



\paragraph{Ethics Statement.}
Our research focuses on evaluating RAG systems through coverage and verifiability assessment and raises no ethical concerns from a methodological standpoint. We collected human annotations via Amazon Mechanical Turk, compensating annotators at \$7.50 per hour, exceeding the U.S.\ federal minimum wage. No personally identifiable information was collected throughout the annotation process. Our experiments use publicly available datasets and well-established LLMs, ensuring compliance with their respective licenses. We therefore confirm that this research raises no ethical concerns. See Appendix~\ref{sec:llm_config.} and Appendix \ref{sec:human_annotation} for the details on LLM configurations and human annotation.

\paragraph{Reproducibility Statement.}
We conducted experiments using both proprietary and open-source LLMs. For proprietary models, we accessed GPT-5, GPT-5-mini, Claude-Sonnet-4, and Gemini-2.5-pro via their respective APIs. For open-source models, we utilized Llama-3.1-8B-Instruct, Llama-3.3-70B-Instruct, Gemma-3-4B-IT, Gemma-3-27B-IT, GPT-oss-20B, Qwen3-4B-Instruct, and Qwen3-30B-A3B-Instruct, all loaded from Hugging Face. Model configurations and evaluation prompts are provided in Appendix~\ref{sec:llm_config.} and Appendix~\ref{sec:prompts}, respectively, ensuring that our experiments are reproducible.

\paragraph{Acknowledgments.}
{This research was supported by the Korea Institute of Science and Technology Information (KISTI) in 2026 (No. (KISTI)K26L3M1C1), aimed at developing KONI (KISTI Open Neural Intelligence), a large language model specialized in science and technology. Additionally, this research was supported by the ``Advanced GPU Utilization Support Program'' funded by the Government of the Republic of Korea (Ministry of Science and ICT) (No. 02-26-01-0181); by the InnoCORE program of the Ministry of Science and ICT (AI Meta-Scientist, N10260110); and by the National Research Foundation of Korea (NRF) funded by Ministry of Science and ICT (RS-2022-NR068758).}

\bibliographystyle{assets/plainnat}
\bibliography{colm2026_conference}

@article{katsis2025mtrag,
author = {Katsis, Yannis and Rosenthal, Sara and Fadnis, Kshitij and Gunasekara, Chulaka and Lee, Young-Suk and Popa, Lucian and Shah, Vraj and Zhu, Huaiyu and Contractor Danish and Danilevsky, Marina},
title = {MTRAG: A Multi-Turn Conversational Benchmark for Evaluating Retrieval-Augmented Generation Systems},
year = {2025},
journal = {arXiv preprint arXiv:2501.03468}
}

@article{li2023electra,
author = {Li, Canjia and Yates, Andrew and MacAvaney, Sean and He, Ben and Sun, Yingfei},
title = {PARADE: Passage Representation Aggregation for Document Reranking},
year = {2023},
journal = {ACM Trans. Inf. Syst.}
}

@article{robertson2009probabilistic,
title={The probabilistic relevance framework: BM25 and beyond},
author={Robertson, Stephen and Zaragoza, Hugo and others},
journal={Foundations and Trends{\textregistered} in Information Retrieval},
volume={3},
number={4},
pages={333--389},
year={2009}
}

@article{zhang2025qwenembed,
title={Qwen3 Embedding: Advancing Text Embedding and Reranking Through Foundation Models},
author={Zhang, Yanzhao and Li, Mingxin and Long, Dingkun and Zhang, Xin and Lin, Huan and Yang, Baosong and Xie, Pengjun and Yang, An and Liu, Dayiheng and Lin, Junyang and Huang, Fei and Zhou, Jingren},
journal={arXiv preprint  arXiv:2506.05176},
year={2025}
}

@article{openai2025gpt5,
  title={GPT-5 System Card},
  author={{OpenAI}},
  year={2025},
  month={August},
}

@article{grattafiori2024llama,
  title={The Llama 3 Herd of Models},
  author={Grattafiori, Aaron and others},
  journal={arXiv preprint arXiv:2407.21783},
  year={2024}
}

@article{yang2025qwen3,
  title={Qwen3 Technical Report},
  author={Yang, An and others},
  journal={arXiv preprint arXiv:2505.09388},
  year={2025}
}

@article{anthropic2025claude,
  title={System Card: Claude Opus 4 \& Claude Sonnet 4},
  author={{Anthropic}},
  year={2025},
}

@article{google2025gemini,
  title={Gemini 2.5: Pushing the Frontier with Advanced Reasoning, Multimodality, Long Context, and Next Generation Agentic Capabilities.},
  author={{Google DeepMind}},
  journal={arXiv preprint arXiv:2507.06261},
  year={2025},
}

@article{google2025gemma,
  title={Gemma 3 Technical Report},
  author={{Google DeepMind}},
  journal={arXiv preprint arXiv:2503.19786},
  year={2025}
}

@article{gan2025ragevalsurvey,
  title={Retrieval Augmented Generation Evaluation in the Era of Large Language Models: A Comprehensive Survey},
  author={Gan, Aoran and Yu, Hao and Liu, Qi and Yan, Wenyu and Huang, Zhenya and Tong, Shiwei and Hu, Guoping},
  journal={arXiv preprint arXiv:2504.14891},
  year={2025}
}

@article{merrick2024arctic,
title={Arctic-Embed: Scalable, Efficient, and Accurate Text Embedding Models}, 
author={Luke Merrick and Danmei Xu and Gaurav Nuti and Daniel Campos},
year={2024},
journal={arXiv preprint arXiv:2405.05374}
}

@article{gao2023retrieval,
  title={Retrieval-augmented generation for large language models: A survey},
  author={Gao, Yunfan and Xiong, Yun and Gao, Xinyu and Jia, Kangxiang and Pan, Jinliu and Bi, Yuxi and Dai, Yi and Sun, Jiawei and Wang, Haofen},
  journal={arXiv preprint arXiv:2312.10997},
  year={2023}
}

@article{kwiatkowski2019natural,
title={Natural Questions: A Benchmark for Question Answering Research},
author={Kwiatkowski, Tom and Palomaki, Jennimaria and Redfield, Olivia and Collins, Michael and Parikh, Ankur and Alberti, Chris and Epstein, Danielle and Polosukhin, Illia and Devlin, Jacob and Lee, Kenton and others},
journal={TACL},
year={2019}
}

@article{yu2024eval,
  title={Evaluation of Retrieval-Augmented Generation: A Survey},
  author={Yu, Hao and Gan, Aoran and Zhang, Kai and Tong, Shiwei and Liu, Qi and Liu, Zhaofeng},
  journal={arXiv preprint arXiv:2405.07437},
  year={2024}
}

@article{wolfson2020break,
title={Break It Down: A Question Understanding Benchmark},
author={Wolfson, Tomer and Geva, Mor and Gupta, Ankit  and Gardner, Matt and Goldberg, Yoav and Deutch, Daniel and Berant, Jonathan},
journal={TACL},
year={2020}
}

@article{jarvelin2002cumulated,
  title={Cumulated Gain-based Evaluation of IR Techniques},
  author={J{\"a}rvelin, Kalervo and Kek{\"a}l{\"a}inen, Jaana},
  journal={ACM Transactions on Information Systems},
  volume={20},
  number={4},
  pages={422--446},
  year={2002}
}

@inproceedings{min2023factscore,
  title={FActScore: Fine-grained Atomic Evaluation of Factual Precision in Long Form Text Generation},
  author={Min, Sewon and Krishna, Kalpesh and Lyu, Xinxi and Lewis, Mike and Yih, Wen-tau and Koh, Pang Wei and Iyyer, Mohit and Zettlemoyer, Luke and Hajishirzi, Hannaneh},
  booktitle={EMNLP},
  year={2023}
}

@inproceedings{wei2024longform,
  title={Long-form Factuality in Large Language Models},
  author={Wei, Jerry and Yang, Chengrun and Song, Xinying and Lu, Yifeng and Hu, Nathan Zixia and Huang, Jie and Tran, Dustin and Peng, Daiyi and Liu, Ruibo and Huang, Da and Du, Cosmo and Le, Quoc V},
  booktitle={NeurIPS},
  year={2024}
}

@inproceedings{manakul2023selfcheckgpt,
  title={SelfCheckGPT: Zero-Resource Black-Box Hallucination Detection for Generative Large Language Models},
  author={Manakul, Potsawee and Liusie, Adian and Gales, Mark J. F.},
  booktitle={EMNLP},
  year={2023}
}

@inproceedings{adam2017newsqa,
  title={NewsQA: A Machine Comprehension Dataset},
  author={Trischler, Adam and Wang, Tong and Yuan, Xingdi and Harris, Justin and Sordoni, Alessandro and Bachman, Philip and Suleman, Kaheer},
  booktitle={ACL},
  year={2017}
}

@inproceedings{jin2019pubmedqa,
  title={PubMedQA: A Dataset for Biomedical Research Question Answering},
  author={Jin, Qiao and Dhingra, Bhuwan and Liu, Zhengping and Cohen, William W. and Lu, Xinghua},
  booktitle={EMNLP},
  year={2019}
}

@inproceedings{yang2018hotpotqa,
title={HotpotQA: A Dataset for Diverse, Explainable Multi-hop Question Answering},
author={Yang, Zhilin and Qi, Peng and Zhang, Saizheng and Bengio, Yoshua and Cohen, William W and Salakhutdinov, Ruslan and Manning, Christopher D},
booktitle={EMNLP},
year={2018}
}

@inproceedings{chen2021finqa,
title={FinQA: A Dataset of Numerical Reasoning over Financial Data},
author={Chen, Zhiyu and Chen, Wenhu and Smiley, Charese and Shah, Sameena and Borova, Iana and Langdon, Dylan and Moussa, Reema and Beane, Matt and Huang, Ting-Hao and Routledge, Bryan and Wang, William Yang},
booktitle={EMNLP},
year={2021}
}

@inproceedings{fan2019eli5,
title={ELI5: Long Form Question Answering},
author={Fan, Angela and Jernite, Yacine and Perez, Ethan and Grangier, David and Weston, Jason and Auli, Michael},
booktitle={ACL},
year={2019}
}

@inproceedings{rau2024bergen,
  title={BERGEN: A Benchmarking Library for Retrieval-Augmented Generation},
  author={Rau, David and Déjean, Hervé and Chirkova, Nadezhda and Formal, Thibault and Wang, Shuai and Nikoulina, Vassilina and Clinchant, Stéphane},
  booktitle={EMNLP},
  year={2024}
}

@inproceedings{es2024ragas,
  title={Ragas: Automated Evaluation of Retrieval Augmented Generation},
  author={Es, Shahul and James, Jithin and Espinosa-Anke, Luis and Schockaert, Steven},
  booktitle={EACL},
  year={2024}
}

@inproceedings{jon2024ares,
  title={ARES: An Automated Evaluation Framework for Retrieval-Augmented Generation Systems},
  author={Saad-Falcon, Jon and Khattab, Omar and Potts, Christopher and Zaharia, Matei},
  booktitle={NAACL},
  year={2024}
}

@inproceedings{Ru2024ragchecker,
  title={RAGChecker: A Fine-grained Framework for Diagnosing Retrieval-Augmented Generation},
  author={Ru, Dongyu and Qiu, Lin and Hu, Xiangkun and Zhang, Tianhang and Shi, Peng and Chang, Shuaichen and Jiayang, Cheng and Wang, Cunxiang and Sun, Shichao and Li, Huanyu and Zhang, Zizhao and Wang, Binjie and Jiang, Jiarong and He, Tong and Wang, Zhiguo and Liu, Pengfei and Zhang, Yue and Zhang, Zheng},
  booktitle={NeurIPS},
  year={2024}
}

@inproceedings{zhu2024rageval,
  title={RAGEval: Scenario Specific RAG Evaluation Dataset Generation Framework},
  author={Zhu, Kunlun and Luo, Yifan and Xu, Dingling and Yan, Yukun and Liu, Zhenghao and Yu, Shi and Wang, Ruobing and Wang, Shuo and Li, Yishan and Zhang, Nan and Han, Xu and Liu, Zhiyuan and Sun, Maosong},
  booktitle={ACL},
  year={2025}
}

@inproceedings{xie2024dorag,
  title={Do RAG Systems Cover What Matters? Evaluating and Optimizing Responses with Sub-Question Coverage},
  author={Xie, Kaige and Laban, Philippe and Choubey, Profulla Kumar and Xiong, Caiming and Wu, Chien-Sheng},
  booktitle={NAACL},
  year={2025}
}

@inproceedings{rassin2024evaluating,
  title={Evaluating D-MERIT of Partial-annotation on Information Retrieval},
  author={Rassin, Royi and Fairstein, Yaron and Kalinsky, Oren and Kushilevitz, Guy and Cohen, Nachshon and Libov, Alexander and Goldberg, Yoav},
  booktitle={EMNLP},
  year={2024}
}

@inproceedings{min2019multi,
title={Multi-hop Reading Comprehension through Question Decomposition and Rescoring},
author={Min, Sewon  and Zhong, Victor  and Zettlemoyer, Luke and Hajishirzi, Hannaneh},
booktitle={ACL},
year={2019}
}

@inproceedings{perez2020unsupervised,
title={Unsupervised Question Decomposition for Question Answering},
author={Perez, Ethan and Lewis, Patrick  and Yih, Wen-tau and Cho, Kyunghyun and Kiela, Douwe},
booktitle={EMNLP},
year={2020}
}

@inproceedings{khot2021text,
  title={Text Modular Networks: Learning to Decompose Tasks in the Language of Existing Models},
  author={Khot, Tushar and Khashabi, Daniel  and Richardson, Kyle  and Clark, Peter and Sabharwal, Ashish },
  booktitle={NAACL},
  year={2021},
}

@inproceedings{press2023selfask,
  title={Measuring and Narrowing the Compositionality Gap in Language Models},
  author={Press, Ofir and Zhang, Muru and Min, Sewon and Schmidt, Ludwig and Smith, Noah A. and Lewis, Mike},
  booktitle={EMNLP},
  year={2023},
}

@inproceedings{shao2023iter,
  title={Enhancing Retrieval-Augmented Large Language Models with Iterative Retrieval-Generation Synergy},
  author={Shao, Zhihong and Gong, Yeyun and Shen, Yelong and Huang, Minlie and Duan, Nan and Chen, Weizhu},
  booktitle={EMNLP},
  year={2023},
}

@inproceedings{rajpurkar2016squad,
  title={SQuAD: 100,000+ Questions for Machine Comprehension of Text},
  author={Rajpurkar, Pranav and Zhang, Jian and Lopyrev, Konstantin and Liang, Percy},
  booktitle={EMNLP},
  year={2016}
}

@inproceedings{lin2004rouge,
  title={ROUGE: A Package for Automatic Evaluation of Summaries},
  author={Lin, Chin-Yew},
  booktitle={Text Summarization Branches Out@ACL},
  year={2004}
}

@inproceedings{salemi2024retqual,
  title={Evaluating Retrieval Quality in Retrieval-Augmented Generation},
  author={Salemi, Alireza and Zamani, Hamed},
  booktitle={SIGIR},
  year={2024},
}

@inproceedings{zhang2024mugi,
title = {Exploring the Best Practices of Query Expansion with Large Language Models},
author = {Zhang, Le  and
Wu, Yihong  and
Yang, Qian  and
Nie, Jian-Yun},
booktitle = {EMNLP},
year = {2024},
}

@inproceedings{yamada2021bpr,
title = {Efficient Passage Retrieval with Hashing for Open-domain Question Answering},
author = {Yamada, Ikuya  and
Asai, Akari  and
Hajishirzi, Hannaneh},
booktitle = {ACL},
year = {2021},
}

@inproceedings{liu2022dimreduction,
title = {Dimension Reduction for Efficient Dense Retrieval via Conditional Autoencoder},
author = {Liu, Zhenghao  and
Zhang, Han  and
Xiong, Chenyan  and
Liu, Zhiyuan  and
Gu, Yu  and
Li, Xiaohua},
booktitle = {EMNLP},
year = "2022"
}

@inproceedings{karpukhin2020dpr,
title = {Dense Passage Retrieval for Open-Domain Question Answering},
author = {Karpukhin, Vladimir and Oguz, Barlas and Min, Sewon and Lewis, Patrick and Wu, Ledell and Edunov, Sergey and Chen, Danqi and Yih, Wen-tau},
booktitle = {EMNLP},
year = {2020},
}

@inproceedings{reimers2019sbert,
title={Sentence-BERT: Sentence Embeddings using Siamese BERT-Networks}, 
author={Nils Reimers and Iryna Gurevych},
year={2019},
booktitle = {EMNLP}
}

@inproceedings{nogueira2020monot5,
title = {Document Ranking with a Pretrained Sequence-to-Sequence Model},
author = {Nogueira, Rodrigo  and
Jiang, Zhiying  and
Pradeep, Ronak  and
Lin, Jimmy},
booktitle = {EMNLP},
year = {2020},
}

@inproceedings{formal2022splade,
author = {Formal, Thibault and Lassance, Carlos and Piwowarski, Benjamin and Clinchant, St\'{e}phane},
title = {From Distillation to Hard Negative Sampling: Making Sparse Neural IR Models More Effective},
year = {2022},
booktitle = {SIGIR}
}

@inproceedings{xiong2021ance,
title={Approximate Nearest Neighbor Negative Contrastive Learning for Dense Text Retrieval}, 
author={Lee Xiong and Chenyan Xiong and Ye Li and Kwok-Fung Tang and Jialin Liu and Paul Bennett and Junaid Ahmed and Arnold Overwijk},
year={2021},
booktitle = {ICLR}
}

@inproceedings{wang2023query2doc,
title={Query2doc: Query Expansion with Large Language Models},
author={Wang, Liang and Yang, Nan and Wei, Furu},
booktitle={EMNLP},
year={2023}
}

@inproceedings{gao2023hyde,
title = {Precise Zero-Shot Dense Retrieval without Relevance Labels},
author = {Gao, Luyu  and
Ma, Xueguang  and
Lin, Jimmy  and
Callan, Jamie},
booktitle = {ACL},
year = {2023}
}

@inproceedings{Takehi2025,
title={LLM-Assisted Relevance Assessments: When Should We Ask LLMs for Help?},
author={Takehi, Rikiya and Voorhees, Ellen M. and Sakai, Tetsuya and Soboroff, Ian},
year={2025},
booktitle={SIGIR}
}

@inproceedings{santhanam2022colbertv2,
title={ColBERTv2: Effective and Efficient Retrieval via Lightweight Late Interaction},
author={Santhanam, Keshav and Khattab, Omar and Saad-Falcon, Jon and Potts, Christopher and Zaharia, Matei},
year={2022},
booktitle={NAACL}
}

@inproceedings{song2024finesure,
  title={FineSurE: Fine-grained Summarization Evaluation using LLMs},
  author={Song, Hwanjun and Su, Hang and Shalyminov, Igor and Cai, Jason and Mansour, Saab},
  booktitle={ACL},
  year={2024}
}

\appendix
\clearpage

\begin{table*}[!t]
\scriptsize
\centering
\setlength{\tabcolsep}{3.2pt}
\begin{tabular}{lllll}
\toprule
Type & Model & Model Checkpoint & Source & Reference \\
\midrule
\multirow{4}{*}{Proprietary} 
& GPT-5-mini & \texttt{gpt-5-mini-2025-08-07} & OpenAI API & \citet{openai2025gpt5} \\
& GPT-5 & \texttt{gpt-5-2025-08-07} & OpenAI API & \citet{openai2025gpt5} \\
& Claude-Sonnet-4 & \texttt{claude-sonnet-4} & Anthropic API & \citet{anthropic2025claude} \\
& Gemini-2.5-pro & \texttt{gemini-2.5-pro} & Google API & \citet{google2025gemini} \\
\midrule
\multirow{8}{*}{Open-source} 
& GPT-oss-20B & \texttt{openai/gpt-oss-20b} & HuggingFace & \citet{openai2025gpt5} \\
& Llama-3.1-8B-Inst. & \texttt{meta-llama/Llama-3.1-8B-Instruct} & HuggingFace & \citet{grattafiori2024llama} \\
& Llama-3.3-70B-Inst. & \texttt{meta-llama/Llama-3.3-70B-Instruct} & HuggingFace & \citet{grattafiori2024llama} \\
& Gemma-3-4B-IT & \texttt{google/gemma-3-4b-it} & HuggingFace & \citet{google2025gemma} \\
& Gemma-3-27B-IT & \texttt{google/gemma-3-27b-it} & HuggingFace & \citet{google2025gemma} \\
& Qwen3-4B-Inst. & \texttt{Qwen/Qwen3-4B-Instruct-2507} & HuggingFace & \citet{yang2025qwen3} \\
& Qwen3-30B-A3B-Inst. & \texttt{Qwen/Qwen3-30B-A3B-Instruct-2507} & HuggingFace & \citet{yang2025qwen3} \\
\bottomrule
\end{tabular}
\vspace{-0.25cm}
\caption{Backbone LLM configurations for answer generation and evaluation with \algname{}.}
\vspace{-0.5cm}
\label{tab:backbone_llm}
\end{table*}

\section{LLM Configurations}
\label{sec:llm_config.}

Table~\ref{tab:backbone_llm} summarizes the LLMs used for answer generation and RAG evaluation using \algname{}. For proprietary models including GPT-5-mini, GPT-5~\citep{openai2025gpt5}, Claude-Sonnet-4~\citep{anthropic2025claude}, and Gemini-2.5-pro~\citep{google2025gemini}, we use their respective APIs with \texttt{temperature=0} for consistent outputs. For open-source models including Llama-3.1-8B-Instruct, Llama-3.3-70B-Instruct~\citep{grattafiori2024llama}, Gemma-3-4B-IT, Gemma-3-27B-IT~\citep{google2025gemma}, GPT-oss-20B~\citep{openai2025gpt5}, Qwen3-4B-Instruct and Qwen3-30B-A3B-Instruct~\citep{yang2025qwen3}, we deploy them locally on NVIDIA RTX PRO 6000 Blackwell Server Edition GPUs (96GB VRAM) using a single GPU per model. All experiments use greedy decoding (\texttt{do\_sample=False}) to ensure reproducibility.

\section{Benchmark Dataset Statistics}
\label{sec:benchmark}
Existing RAG benchmarks lack comprehensive coverage of both close-ended and open-ended queries for evaluating coverage and verifiability. To address this gap, we construct a benchmark spanning diverse domains with human annotations. Table~\ref{tab:dataset} presents the detailed statistics, and the annotation process is described in Appendix \ref{sec:human_annotation}.

\smallskip\smallskip
\noindent\textbf{Close-ended Datasets.}
Close-ended queries expect a single definitive answer. We include datasets from general, news, and finance domains:
\emph{NQ}~\citep{kwiatkowski2019natural} (Natural Questions) contains real user queries from Google Search with answers from Wikipedia;
\emph{HotpotQA}~\citep{yang2018hotpotqa} requires multi-hop reasoning over multiple Wikipedia paragraphs;
\emph{NewsQA}~\citep{adam2017newsqa} consists of reading comprehension questions over CNN news articles;
\emph{FinQA}~\citep{chen2021finqa} involves numerical reasoning over financial reports.

\smallskip\smallskip
\noindent\textbf{Open-ended Datasets.}
Open-ended queries allow multiple valid answers and typically require long-form explanations addressing multiple aspects of the query. We include datasets from general, biomedical, science, and technology domains:
\emph{PubMedQA}~\citep{jin2019pubmedqa} comprises biomedical questions derived from PubMed abstracts;
\emph{LoTTE-Science} and \emph{LoTTE-Technology}~\citep{santhanam2022colbertv2} contain practical questions from StackExchange forums in scientific and technical domains, respectively; \emph{ELI5}~\citep{fan2019eli5} (Explain Like I'm Five) consists of questions from Reddit.

\smallskip\smallskip
\noindent\textbf{Statistics.}
The benchmark comprises 320 queries (160 per query type) with human annotations for sub-query--chunk, sub-query--claim, and claim--chunk alignments. 
The statistics reveal distinct patterns between query types. 
Close-ended queries yield fewer sub-queries on average (1.51) and claims (1.50), while open-ended queries produce more sub-queries (2.04) and substantially more claims (4.69), reflecting their inherent complexity requiring multi-aspect explanations.
Notably, within close-ended datasets, multi-hop reasoning datasets (HotpotQA, FinQA) yield significantly more sub-queries (1.57--2.17) than single-focus datasets such as NQ and NewsQA (1.1). This decomposition enables fine-grained query coverage evaluation by assessing how well each retrieved chunk addresses individual reasoning steps.
This benchmark enables evaluation of how well RAG metrics correlate with human judgments across different query types and domains.

\begin{table*}[!t]
\centering
\scriptsize
\setlength{\tabcolsep}{1.7pt}
\begin{tabular}{llllrccc cc}
\toprule
Category & Dataset & Domain & Source & Corpus Size & \#Query & \#Sub-query & \#Claim & Avg Sub-query & Avg Claim \\
\midrule
\multirow{5}{*}{Close-ended} 
 & NQ & General & Wikipedia & 5,863,011 & 40 & 44 & 434 & 1.10 & 1.36 \\
 & HotpotQA & General & Wikipedia & 5,863,011 & 40 & 63 & 410 & 1.57 & 1.28 \\
 & NewsQA & News & CNN articles & 117,559 & 40 & 47 & 616 & 1.18 & 1.93 \\
 & FinQA & Finance & Financial reports & 50,018 & 40 & 87 & 458 & 2.17 & 1.50 \\
\cmidrule{2-10}
 & \multicolumn{4}{l}{\textit{Subtotal}} & 160 & 241 & 1,918 & 1.51 & 1.50 \\
\midrule
\multirow{5}{*}{Open-ended} 
 & PubMedQA & Biomedical & PubMed & 651,259 & 40 & 74 & 1,166 & 1.85 & 3.64 \\
 & LoTTE-Sci. & Science & StackExchange & 30,452 & 40 & 79 & 1,758 & 1.98 & 5.53 \\
 & LoTTE-Tech. & Technology & StackExchange & 30,452 & 40 & 65 & 1,513 & 1.62 & 4.73 \\
 & ELI5 & General & Reddit & 1,656,096 & 100 & 108 & 1,552 & 2.70 & 4.85 \\
\cmidrule{2-10}
 & \multicolumn{4}{l}{\textit{Subtotal}} & 160 & 326 & 5,989 & 2.04 & 4.69 \\
\bottomrule
\end{tabular}
\vspace{-0.25cm}
\caption{Benchmark dataset statistics. \#Subquery and \#Claim denote the total number of decomposed sub-queries and answer claims, respectively. Avg Sub and Avg Fact represent the average per query. For each query, we retrieve top-10 chunks using BM25.}
\vspace{-0.5cm}
\label{tab:dataset}
\end{table*}

\section{Human Annotation}
\label{sec:human_annotation}

\smallskip\smallskip
\noindent\textbf{Recruitment and Compensation.}
We recruit annotators through Amazon Mechanical Turk (MTurk) with strict qualification criteria to ensure annotation quality, including a HIT approval rate above 90\%, at least 500 approved HITs, and a minimum score of 90 on our custom English comprehension exam. Annotators are compensated \$7.5 per hour, exceeding the U.S. federal minimum wage.

\smallskip\smallskip
\noindent\textbf{Instructions and Consent.}
Annotators are provided with detailed instructions adapted from the LLM prompts in Appendix \ref{sec:prompts}, rewritten in a human-readable format. The instructions clearly explain that the collected annotations will be used solely for research purposes and may be publicly released as part of an academic benchmark. By accepting the task, annotators provide informed consent to these terms. The annotation task involves only objective labeling (e.g., verifying decomposition validity and pairwise alignment) and does not collect any personal information or opinions from annotators.

\smallskip\smallskip
\noindent\textbf{Annotation Process.}
We collect human annotations for two types of tasks. First, for decomposition quality, annotators verify whether GPT-5 generated sub-queries and answer claims are valid. As shown in Table~\ref{tab:agreement} in the main paper, over 97\% of GPT-5 decomposition results are approved by all three annotators, confirming the reliability of LLM-based decomposition. Second, for alignment checks, annotators perform pairwise alignment labeling for sub-query--chunk, sub-query--claim, and claim--chunk pairs. Each instance is evaluated by three qualified annotators, and final labels are determined by majority vote. We insert 10\% attention checks into each HIT, and annotators must pass all attention checks for their responses to be accepted. This strict filtering improves overall consistency.

\section{Details of Conventional Metrics}
\label{sec:detail_conventional}
In this section, we summarize the conventional metrics used for generator evaluation (Section \ref{sec:impact_coverage_aware}). Conventional metrics such as Exact Match, Accuracy, and ROUGE-L are lexical-based metrics that evaluate generators by measuring word overlap between generated and reference answers~\citep{lin2004rouge, rajpurkar2016squad, gan2025ragevalsurvey}. For instance, given the query ``What is the capital of France?'', generated answer ``Paris'', and reference answer ``The capital of France is Paris'', Exact Match evaluates it as incorrect since the strings do not match exactly, while Accuracy evaluates it as correct since the generated answer is contained in the reference, and ROUGE-L yields a partial score of 0.29. These metrics are defined as follows:

\smallskip\smallskip
\noindent\textbf{Exact Match (EM).}
This metric measures whether the generated answer $\hat{y}$ is identical to the reference answer $y$:
\begin{equation}
    \text{EM} = \mathbf{1}[\hat{y} = y],
\end{equation}
where $\mathbf{1}[\cdot]$ is the indicator function that returns 1 if the condition is satisfied and 0 otherwise. Both answers are normalized (lowercasing, removing punctuation and articles) before comparison. EM strictly returns 1 only when the generated answer matches the reference answer exactly.

\smallskip\smallskip
\noindent\textbf{Accuracy (ACC).}
As a relaxed version of EM, this metric checks whether the generated answer $\hat{y}$ is contained within the reference answer $y$ or vice versa:
\begin{equation}
    \text{ACC} = \mathbf{1}[\hat{y} \subseteq y \lor y \subseteq \hat{y}],
\end{equation}
where $\subseteq$ denotes substring inclusion and $\lor$ denotes logical OR. Both answers are normalized before comparison. This metric accommodates cases where answers differ in verbosity but contain equivalent information.

\smallskip\smallskip
\noindent\textbf{ROUGE-L.}
This metric measures the longest common subsequence (LCS) between the generated answer $\hat{y}$ and reference answer $y$:
\begin{equation}
    R_{\text{lcs}} = \frac{|\text{LCS}(\hat{y}, y)|}{|y|}, \quad P_{\text{lcs}} = \frac{|\text{LCS}(\hat{y}, y)|}{|\hat{y}|},
\end{equation}
\begin{equation}
    \text{ROUGE-L} = \frac{(1 + \beta^2) \cdot P_{\text{lcs}} \cdot R_{\text{lcs}}}{R_{\text{lcs}} + \beta^2 \cdot P_{\text{lcs}}},
\end{equation}
where $|\cdot|$ denotes the number of tokens and $\beta$ is typically set as hyperparameter.


\section{Latency}
\label{sec:latency}

\smallskip\smallskip
\noindent\textbf{Setup.}
We measure average latency per query for all methods on one or two NVIDIA RTX PRO 6000 Blackwell Server Edition GPUs. Table~\ref{tab:latency} reports the latency comparison with other LLM-based metrics, and Table~\ref{tab:backbone_latency} reports the latency of \algname{} with different backbone LLMs.

\begin{wraptable}{r}{0.62\textwidth}
\centering
\scriptsize
\setlength{\tabcolsep}{2.8pt}
\vspace*{-0.4cm}
\begin{tabular}{lrr}
\toprule
Method / Step & Close-ended & Open-ended \\
\midrule
RAGEval    & 21.85s  & 50.22s  \\
RAGAS      & 69.78s  & 152.25s \\
RAGChecker & 35.89s  & 137.54s \\
DoRAG      & 389.63s & 466.59s \\
\midrule
\textbf{\algname{} (Total)} & \textbf{54.65s} & \textbf{97.71s} \\
\rowcolor{gray!10} \quad Query Decomposition (\S\ref{subsec:decomposition}) & 7.07s & 7.14s \\
\quad Answer Decomposition (\S\ref{subsec:decomposition}) & 3.53s & 21.10s \\
\rowcolor{gray!10} \quad Sub-query--Chunk Alignment (Eq.\eqref{eq:query-chunk_align}) & 21.47s & 22.51s \\
\quad Sub-query--Claim Alignment (Eq.\eqref{eq:query-claim_align}) & 2.57s & 18.56s \\
\rowcolor{gray!10} \quad Claim--Chunk Alignment (Eq.\eqref{eq:claim-chunk_align}) & 18.80s & 26.61s \\
\quad Query Coverage Check & 1.21s & 1.79s \\
\bottomrule
\end{tabular}
\vspace{-0.15cm}
\caption{Average latency per query of LLM-based metrics with step-by-step breakdown for \algname{}.}
\label{tab:latency}
\vspace*{-0.45cm}
\end{wraptable}

\smallskip\smallskip
\noindent\textbf{Comparison with LLM-based Metrics.}
We compare \algname{} against other LLM-based metrics using Qwen3-30B-A3B-Inst. as the backbone. Among the LLM-based metrics, G-Eval is the fastest as it relies on a simple prompt-based evaluation, while DoRAG is the slowest due to its 20 sub-question generation and classification process. \algname{} achieves moderate latency (54.65s for close-ended, 97.71s for open-ended), comparable to RAGChecker and significantly faster than DoRAG. We also provide a breakdown of \algname{}'s latency by step. Query decomposition and answer decomposition are relatively fast, while the chunk-related alignment steps (Claim--Chunk and Sub-query--Chunk) account for the largest portion, as they require pairwise comparisons across all retrieved chunks. Sub-query--Claim alignment is comparatively fast since it only involves alignment within the decomposed elements. Open-ended queries take longer overall due to the larger number of sub-queries and answer claims generated during decomposition. Note that alignment checks can be parallelized across chunks, offering potential for further latency reduction in practice.

\smallskip\smallskip
\noindent\textbf{Comparison across Backbone Models.}
Table \ref{tab:backbone_latency} compares \algname{}'s latency across different backbone LLMs. Smaller models such as Llama-3.1-8B-Inst. are faster but exhibit lower correlation with human judgments (Table~\ref{tab:backbone} in Section \ref{sec:baseline}). 
\begin{wraptable}{r}{0.45\textwidth}
\centering
\scriptsize
\setlength{\tabcolsep}{2pt}
\vspace*{-0.4cm}
\begin{tabular}{lrr}
\toprule
Backbone Model & Close-ended (s) & Open-ended (s) \\
\midrule
Llama-3.1-8B-Inst.  & 34.59s & 29.20s \\
Llama-3.3-70B-Inst. & 51.38s & 52.96s \\
Qwen3-30B-A3B-Inst. & 54.65s & 97.71s \\
\bottomrule
\end{tabular}
\vspace{-0.25cm}
\caption{Average latency per query of \algname{} with different backbone LLMs.}
\label{tab:backbone_latency}
\vspace*{-0.45cm}
\end{wraptable}Larger models such Llama-3.3-70B-Inst. and Qwen3-30B-A3B-Inst. require moderately longer evaluation time, yet remain significantly faster than DoRAG (389.63s and 466.59s for close-ended and open-ended, respectively) while achieving higher correlation with human judgments. This demonstrates that \algname{} offers a favorable trade-off between evaluation quality and computational cost across various backbone model sizes.

\begin{wraptable}{r}{0.5\textwidth}
\centering
\scriptsize
\setlength{\tabcolsep}{2.7pt}
\begin{tabular}{lccccr}
\toprule
\multirow{2}{*}{Method} & \multicolumn{4}{c}{Pearson Correlation} & \multirow{2}{*}{Latency} \\
\cmidrule(lr){2-5}
 & Prec. & Comp. & Conc. & Veri. & \\
\midrule
\multicolumn{6}{l}{\textit{Close-ended}} \\
\algname{}                      & 0.55 & 0.52 & 0.47 & 0.40 & 54.7s \\
\texttt{Q-CARE-Lite} (Chunk 2)  & 0.61 & 0.56 & 0.49 & 0.40 & 45.3s \\
\texttt{Q-CARE-Lite} (Chunk 5)  & 0.35 & 0.54 & 0.49 & 0.29 & 25.9s \\
\texttt{Q-CARE-Lite} (Chunk 10) & 0.19 & 0.54 & 0.48 & 0.38 & 20.1s \\
\midrule
\multicolumn{6}{l}{\textit{Open-ended}} \\
\algname{}                      & 0.49 & 0.50 & 0.38 & 0.69 & 97.7s \\
\texttt{Q-CARE-Lite} (Chunk 2)  & 0.51 & 0.49 & 0.46 & 0.62 & 88.6s \\
\texttt{Q-CARE-Lite} (Chunk 5)  & 0.36 & 0.47 & 0.42 & 0.58 & 68.2s \\
\texttt{Q-CARE-Lite} (Chunk 10) & 0.38 & 0.50 & 0.45 & 0.52 & 57.1s \\
\bottomrule
\end{tabular}
\vspace{-0.25cm}
\caption{Latency and Pearson correlation with human judgments for \texttt{Q-CARE-Lite} across chunk batch sizes. Prec.: C-Prec@10, Comp.: Completeness, Conc.: Conciseness, Veri.: Verifiableness.}
\label{tab:lite}
\vspace*{-0.45cm}
\end{wraptable}

\smallskip\smallskip
\noindent\textbf{Latency Reduction via Chunk Batching.}
To further reduce latency for practical use, we propose \texttt{Q-CARE-Lite}, which batches multiple chunks into a single LLM inference call, substantially reducing the total number of LLM calls required for alignment checks. We denote the batch size as ``Chunk $N$'', where $N$ chunks are evaluated together in one call. Table~\ref{tab:lite} reports both the estimated latency and the quality tradeoff across batch sizes. \texttt{Q-CARE-Lite} (Chunk 2) achieves comparable quality across most dimensions while reducing close-ended latency by $\sim$17\%. For more aggressive latency reduction, Chunk 5 cuts latency by $\sim$30--53\% with moderate quality tradeoffs mainly in Precision. We recommend \texttt{Q-CARE-Lite} (Chunk 2) as a lightweight configuration that balances evaluation quality and computational efficiency.

\section{Definition of Composite Score}
\label{sec:composite}
To compute correlations between human judgments and evaluation metrics, we aggregate the generator evaluation metrics computed from human annotations in Section~\ref{subsec:metric} into a single composite score. This aggregation is necessary for two reasons. First, traditional metrics such as EM, ACC, and ROUGE-L produce a single scalar value per query-answer pair, making direct comparison with multi-dimensional evaluations infeasible. Second, measuring the correlation between retriever performance and generation quality requires an overall assessment of the generated answer; without a unified score, it is difficult to quantify how retrieval quality impacts downstream generation across multiple evaluation dimensions from a holistic perspective.

We define the human composite score as the  mean of three evaluation generator metrics:

\begin{equation}
\text{Composite} = \frac{\text{Comp.} + \text{Conc.} + \text{Veri.}}{3}
\end{equation}

where Comp., Conc., and Veri. denote Completeness, Conciseness, and Verifiableness, respectively. By averaging across these generator metrics, the composite score captures overall answer quality while balancing potential trade-offs, such as the tension between completeness and conciseness. We use this composite score for computing correlations in Section \ref{sec:impact_coverage_aware}.

\section{Details of Benchmarking Configuration}
\label{sec:benchmark_config}
In this section, we provide details on the benchmarking configuration for evaluating \algname{}. We describe the query sampling procedure, the retriever configurations, and the generator models used in our experiments.

\smallskip\smallskip
\noindent\textbf{Query Sampling.}
We construct our evaluation set by randomly sampling queries from each dataset. For close-ended datasets (NQ, HotpotQA, NewsQA, FinQA), we randomly sample 100 queries per dataset, yielding 400 close-ended queries in total. Similarly, for open-ended datasets (PubMedQA, LoTTE-Science, LoTTE-Technology, ELI5), we randomly sample 100 queries per dataset, yielding 400 open-ended queries. This results in a total of 800 queries across both query types. Random sampling ensures diverse coverage of query characteristics within each dataset while maintaining balanced representation across domains.

\smallskip\smallskip
\noindent\textbf{Retriever Configurations.}
Table~\ref{tab:retriever_config} summarizes the retriever configurations used for benchmarking. By combining base retrievers with rerankers and query expansion methods, we benchmark a total of 21 retriever configurations.
\begin{itemize}[leftmargin=*, noitemsep, topsep=2pt]
\item \textbf{Sparse Retrievers}: These methods represent queries and documents as high-dimensional sparse vectors based on lexical features. BM25~\citep{robertson2009probabilistic} is a classic lexical matching method based on term frequency and inverse document frequency. SPLADE~\citep{formal2022splade} learns sparse representations through expansion, combining the interpretability of sparse methods with learned representations.

\item \textbf{Dense Retrievers}: These methods encode queries and documents into dense vector representations, enabling semantic similarity matching. SBERT~\citep{reimers2019sbert} generates sentence embeddings for semantic similarity matching. DPR~\citep{karpukhin2020dpr} is trained with contrastive learning on question-passage pairs. ANCE~\citep{xiong2021ance} improves training with hard negatives sampled from the retriever itself. BPR~\citep{yamada2021bpr} incorporates entity information for better retrieval. Dim Reduction~\citep{liu2022dimreduction} applies dimensionality reduction techniques for efficient retrieval. Arctic~\citep{merrick2024arctic} and Qwen3-Embedding~\citep{zhang2025qwenembed} \{0.6B, 4B, 8B\} represent recent advances in embedding quality.

\item \textbf{Rerankers}: These methods re-score an initial set of retrieved candidates to improve ranking quality. MonoT5~\citep{nogueira2020monot5} uses a sequence-to-sequence model to score query-document relevance. ELECTRA~\citep{li2023electra} applies a discriminative model for reranking. Both rerankers re-score candidates initially retrieved by BM25 or ANCE.

\begin{figure*}[!t]
\centering
\includegraphics[width=14cm]{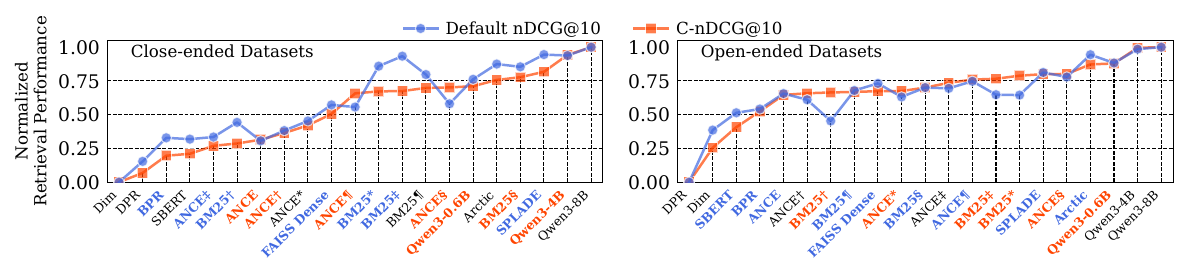}
\vspace*{-1cm}
\caption{Retriever performance with Default nDCG@10 and Coverage-aware nDCG@10, sorted by the latter. Retriever names are colored {\color[HTML]{4169E1}blue} if ranked higher by Default nDCG@10, {\color[HTML]{FF4500}orange} if ranked higher by Coverage-aware nDCG@10. Superscripts denote query expansion (\textsuperscript{\textdagger}HyDE, \textsuperscript{\textdaggerdbl}Q2D, \textsuperscript{*}MuGI) and reranking (\textsuperscript{\textsection}monoT5, \textsuperscript{\textparagraph}ELECTRA).}
\label{fig:retriever_ndcg}
\vspace*{-0.3cm}
\end{figure*}

\begin{table*}[!t]
\centering
\scriptsize
\setlength{\tabcolsep}{4pt}
\begin{tabular}{lllp{7cm}}  
\toprule
Type & Model & Reference & Source \\
\midrule
\multirow{2}{*}{Sparse}
& BM25   & \citet{robertson2009probabilistic} & \url{https://pypi.org/project/pyserini/} \\
& SPLADE & \citet{formal2022splade}            & \url{https://huggingface.co/naver/splade-cocondenser-\-ensembledistil} \\
\midrule
\multirow{10}{*}{Dense}
& SBERT          & \citet{reimers2019sbert}    & \url{https://huggingface.co/sentence-transformers/msmarco-distilbert-base-tas-b} \\
& DPR            & \citet{karpukhin2020dpr}    & \url{https://huggingface.co/sentence-transformers/facebook-dpr-ctx_encoder-multiset-base} \\
& ANCE           & \citet{xiong2021ance}       & \url{https://huggingface.co/sentence-transformers/msmarco-roberta-base-ance-firstp} \\
& BPR            & \citet{yamada2021bpr}       & \url{https://huggingface.co/income/bpr-gpl-trec-\-covid-base-msmarco-distilbert-tas-b} \\
& Arctic         & \citet{merrick2024arctic}   & \url{https://huggingface.co/Snowflake/snowflake-arctic-embed-m} \\
& Dim Reduction  & \citet{liu2022dimreduction} & \url{https://huggingface.co/sentence-transformers/msmarco-distilbert-base-tas-b} \\
& Faiss Dense    & \citet{karpukhin2020dpr}    & \url{https://github.com/facebookresearch/faiss/wiki} \\
& Qwen3-0.6B     & \citet{zhang2025qwenembed}  & \url{https://huggingface.co/Qwen/Qwen3-Embedding-0.6B} \\
& Qwen3-4B       & \citet{zhang2025qwenembed}  & \url{https://huggingface.co/Qwen/Qwen3-Embedding-4B} \\
& Qwen3-8B       & \citet{zhang2025qwenembed}  & \url{https://huggingface.co/Qwen/Qwen3-Embedding-8B} \\
\midrule
\multirow{2}{*}{Reranker}
& MonoT5  & \citet{nogueira2020monot5} & \url{https://huggingface.co/castorini/monot5-base-msmarco} \\
& ELECTRA & \citet{li2023electra}      & \url{https://huggingface.co/cross-encoder/ms-marco-electra-base} \\
\midrule
\multirow{3}{*}{\makecell[l]{Query\\Expansion}}
& HyDE & \citet{gao2023hyde}      & \url{https://github.com/texttron/hyde} \\
& Q2D  & \citet{wang2023query2doc} & \url{https://arxiv.org/pdf/2303.07678} \\
& MuGI & \citet{zhang2024mugi}     & \url{https://github.com/lezhang7/Retrieval_MuGI} \\
\bottomrule
\end{tabular}
\vspace{-0.3cm}
\caption{Retriever configurations used for benchmarking in Section \ref{benchmarking}.}
\vspace{-0.5cm}
\label{tab:retriever_config}
\end{table*}

\item \textbf{Query Expansion Methods}: These methods enhance the original query by generating additional terms or pseudo-documents to improve retrieval coverage. HyDE~\citep{gao2023hyde} generates a hypothetical document from the query and retrieves using its embedding. Q2D~\citep{wang2023query2doc} expands queries with LLM-generated pseudo-documents. MuGI~\citep{zhang2024mugi} uses multiple generated queries for diverse retrieval. These methods are combined with BM25 or ANCE as the base retriever.
\end{itemize}
By combining base retrievers with rerankers and query expansion methods, we benchmark a total of 21 retriever configurations. All configurations retrieve top-10 chunks per query for evaluation. As shown in Figure~\ref{fig:retriever_ndcg}, comparing Default nDCG@10 and C-nDCG@10 reveals notable ranking discrepancies: 14 retrievers change ranks on close-ended datasets and 15 on open-ended datasets. This trend is consistent with the C-Prec@10 results reported in the main paper (Figure~\ref{fig:retriever_precision}), further validating that coverage-aware metrics provide complementary insights into retrieval quality.

\smallskip\smallskip
\noindent\textbf{Generator Configurations.}
Table~\ref{tab:backbone_llm} incorporates the generator models used for answer generation. We select eight models spanning both proprietary and open-source categories to ensure diverse evaluation:

\begin{itemize}[leftmargin=*, noitemsep, topsep=2pt]
\item \textbf{Proprietary Models}: GPT-5~\citep{openai2025gpt5} represents the state-of-the-art from OpenAI. Claude-Sonnet-4~\citep{anthropic2025claude} is Anthropic's balanced model optimized for both capability and efficiency. Gemini-2.5-pro~\citep{google2025gemini} is Google's flagship model.

\item \textbf{Open-source Models}: GPT-oss-20B~\citep{openai2025gpt5} is OpenAI's open-source release. Qwen3-4B-Inst. and Qwen3-30B-A3B-Inst.~\citep{yang2025qwen3} represent small and medium-scale models from the Qwen family, with the latter using mixture-of-experts architecture. Gemma-3-4B-IT and Gemma-3-27B-IT~\citep{google2025gemma} are Google's open-weight instruction-tuned models at different scales.
\end{itemize}

This selection covers a range of model sizes, and sources (proprietary and open-source), enabling comprehensive evaluation of generator quality across different resource constraints.


\section{Stability in \algname{} Evaluation}

\label{sec:stability}

Since LLM-based evaluation methods may produce varying outputs across runs due to inherent stochasticity, it is important to verify that evaluation scores remain stable. In this section, we assess the consistency of \algname{} across multiple runs.

\begin{wraptable}{r}{0.54\textwidth}
\centering
\scriptsize
\setlength{\tabcolsep}{0.5pt}
\vspace*{-0.4cm}
\begin{tabular}{lccccccccccc}
\toprule
\multirow{2}{*}{Method} & \multicolumn{5}{c}{Close-ended} & \multicolumn{5}{c}{Open-ended} \\
\cmidrule(lr){2-6} \cmidrule(lr){7-11}
 & Comp. & Conc. & Veri. & Prec. & nDCG. & Comp. & Conc. & Veri. & Prec. & nDCG. \\
\midrule
\algname{} & 0.83 & 0.84 & 0.91 & 0.93 & 0.86 & 0.79 & 0.83 & 0.94 & 0.90 & 0.89 \\
\bottomrule
\end{tabular}
\vspace{-0.15cm}
\caption{Inter-annotator agreement (Krippendorff's alpha) of \algname{} across three independent runs.}
\label{tab:stability}
\vspace*{-0.45cm}
\end{wraptable}

\smallskip\smallskip
\noindent\textbf{Setup.} We run the evaluation pipeline three times with Qwen3-30B-A3B-Instruct and measure inter-annotator agreement (IAA) via Krippendorff's alpha, treating each independent run as a separate annotator. Low variability across runs leads to high IAA, directly indicating stable evaluation.

\smallskip\smallskip
\noindent\textbf{Results.} Table~\ref{tab:stability} reports the results for each evaluation dimension. All dimensions achieve IAA above 0.79, with Verifiableness and Precision exceeding 0.90 on both query types. According to standard interpretation guidelines, alpha values above 0.80 indicate strong agreement, confirming that \algname{} produces reliable and reproducible evaluations. Notably, the high consistency is maintained across both close-ended and open-ended queries, demonstrating that \algname{}'s query-answer decomposition and alignment check stages yield stable results regardless of query types. These results suggest that \algname{} can serve as a dependable evaluation framework for comparing RAG systems across multiple runs without significant variance.

\section{Baseline Metric Mapping}
\label{sec:baseline_mertic_mapping}



\begin{wraptable}{r}{0.54\textwidth}
\centering
\scriptsize
\setlength{\tabcolsep}{0.8pt}
\vspace*{-0.4cm}
\begin{tabular}{lcccc}
\toprule
\algname{} & Completeness & Conciseness & Verifiableness & Precision@10 \\
\midrule
RAGEval & Completeness & N/A & \shortstack{1-Halluci-\\nation} & EIR \\
\rowcolor{gray!10} RAGAs & \shortstack{Answer-\\Relevance} & N/A & Faithfulness & \shortstack{Context-\\Relevance} \\
RAGChecker & \shortstack{Claim-\\Recall} & Precision & Faithfulness & \shortstack{Context-\\Precision} \\
\rowcolor{gray!10} DoRAG & \shortstack{Core-\\coverage} & N/A & N/A & N/A \\
\bottomrule
\end{tabular}
\vspace{-0.15cm}
\caption{Metric mapping between baseline methods and \algname{} evaluation dimensions.}
\label{tab:metric_mapping}
\vspace*{-0.45cm}
\end{wraptable}

To ensure fair comparison, we carefully map each baseline's metrics to our evaluation dimensions based on their underlying objectives. Table~\ref{tab:metric_mapping} summarizes the mapping.

\noindent$\bullet$~\emph{RAGAs}: Answer-Relevance measures how well the answer addresses the query (Completeness). Faithfulness assesses whether claims are grounded in retrieved context (Verifiableness). Context-Relevance evaluates retrieval coverage (Precision@10).

\noindent$\bullet$~\emph{RAGChecker}: Claim-Recall measures the proportion of reference claims covered by the answer (Completeness). Precision captures the proportion of answer claims that are relevant (Conciseness). Faithfulness measures factual grounding in retrieved chunks (Verifiableness). Context-Precision evaluates retrieval precision (Precision@10).

\noindent$\bullet$~\emph{RAGEval}: We use its Completeness metric directly. 1-Hallucination measures factual grounding (Verifiableness). EIR (Effective Information Rate) quantifies the proportion of relevant information within retrieved passages (Precision@10).

\noindent$\bullet$~\emph{DoRAG}: It defines multiple metrics for sub-question coverage analysis, but only Core-coverage, which corresponds to Metric \#1 (core) in \citet{xie2024dorag}, aligns with our dimensions (Completeness).

Most baselines lack a metric corresponding to Conciseness, as they primarily focus on whether the query is answered and the answer is faithful. RAGChecker provides the most comprehensive matching among the baselines, covering all four evaluation dimensions.


\begin{table}[!t]
\centering
\scriptsize
\setlength{\tabcolsep}{3.5pt}
\begin{tabular}{@{}p{0.98\columnwidth}@{}}
\toprule
\textbf{Query} \\
\midrule
$\mathcal{Q}$: Prostate-specific antigen and free prostate-specific antigen in the early detection of prostate cancer: do combination tests improve detection? \\
\midrule
\textbf{Model Response} \\
\midrule
$\mathcal{A}$: Combination tests using free and total prostate-specific antigen (PSA), such as the Prostate Health Index, can improve detection, but results have been controversial. Some tests, like \%p2PSA and the Prostate Health Index, were more accurate predictors of disease and increased the accuracy of ... \\
\midrule
\textbf{Decomposed Sub-queries} \\
\midrule
$q_1$: What is the diagnostic performance of total prostate-specific antigen (PSA) testing alone for early detection of prostate cancer? \\
$q_2$: What is the diagnostic performance of free prostate-specific antigen (free PSA) testing alone for early detection of prostate cancer? \\
$q_3$: Does combining total PSA and free PSA into a single testing strategy improve early detection of prostate? \\
\midrule
\textbf{Extracted Claims} \\
\midrule
$a_1$: Combination tests using free and total prostate-specific antigen (PSA), such as the Prostate Health Index, can improve detection. \\
$a_2$: Results evaluating combination tests using free and total prostate-specific antigen (PSA) have been controversial. \\
$a_3$: \%p2PSA and the Prostate Health Index were more accurate predictors. \\
$a_4$: \%p2PSA and the Prostate Health Index increased the accuracy of models including prostate-specific antigen (PSA) and prostate volume. \\
$a_5$: The Prostate Health Index outperformed some tests.
$f_6$: The Prostate Health Index did not outperform free PSA or the free/total PSA ratio.\\
\bottomrule
\end{tabular}
\vspace{-0.2cm}
\caption{Decomposition example from PubMedQA.}
\vspace{-0.1cm}
\label{tab:decomposition_example}
\end{table}

\section{Analysis of 3-level Alignment}
\label{sec:graded}
While \algname{} adopts a binary Yes/No alignment scheme for simplicity and reproducibility, the underlying framework naturally generalizes to finer-grained scales. To explore this flexibility, we extend the alignment judgments to a three-level scale (Yes = 1, Partial = 0.5 and No = 0) for two core alignment steps: (a) sub-query $\leftrightarrow$ answer coverage and (b) sub-query $\leftrightarrow$ chunk coverage, and examine how 3-level evidence distributes across RAG systems.

\begin{wraptable}{r}{0.52\textwidth}
\centering
\scriptsize
\setlength{\tabcolsep}{9pt}
\vspace*{-0.4cm}
\begin{tabular}{llcc}
\toprule
Alignment & Label & Close-ended & Open-ended \\
\midrule
\multirow{3}{*}{\shortstack[l]{Sub-query\\$\leftrightarrow$ Claim}}
 & Yes     & 57.7 & 76.3 \\
 & Partial & 3.9  & 6.6  \\
 & No      & 39.4 & 17.2 \\
\midrule
\multirow{3}{*}{\shortstack[l]{Sub-query\\$\leftrightarrow$ Chunk}}
 & Yes     & 24.4 & 14.7 \\
 & Partial & 11.3 & 17.1 \\
 & No      & 64.3 & 68.4 \\
\bottomrule
\end{tabular}
\vspace{-0.15cm}
\caption{Average distribution (\%) of Yes / Partial / No judgments under 3-level alignment.}
\label{tab:graded_dist}
\vspace*{-0.45cm}
\end{wraptable}

\smallskip\smallskip
\noindent\textbf{Distribution of 3-level Judgments.}
Table~\ref{tab:graded_dist} reports the proportion of Yes, Partial, and No judgments across all eight evaluated RAG systems, revealing two consistent patterns. First, claim-level Partial alignments are sparse ($\sim$2--7\% for close-ended and $\sim$4--10\% for open-ended), suggesting that answer claims tend to either clearly address a sub-query or not, validating the binary assumption at the answer level. Second, chunk-level Partial alignments are more prevalent ($\sim$11\% for close-ended and $\sim$17\% for open-ended), indicating that a non-trivial fraction of sub-query--chunk pairs involve implicit or borderline evidence, which a 3-level scale can more faithfully capture.

\begin{wraptable}{r}{0.52\textwidth}
\centering
\scriptsize
\setlength{\tabcolsep}{7.2pt}
\vspace*{-0.4cm}
\begin{tabular}{lcccc}
\toprule
\multirow{2}{*}{Metric} & \multicolumn{2}{c}{Close-ended} & \multicolumn{2}{c}{Open-ended} \\
\cmidrule(lr){2-3} \cmidrule(lr){4-5}
 & Binary & 3-Level & Binary & 3-Level \\
\midrule
C-Precision@10 & 0.55 & 0.67 & 0.49 & 0.47 \\
Completeness   & 0.52 & 0.52 & 0.50 & 0.48 \\
Conciseness    & 0.47 & 0.47 & 0.38 & 0.42 \\
Verifiableness & 0.40 & 0.47 & 0.69 & 0.72 \\
\bottomrule
\end{tabular}
\vspace{-0.15cm}
\caption{Pearson correlation with human judgments under binary vs.\ 3-level alignment. Prec.: C-Prec@10, Comp.: Completeness, Conc.: Conciseness, Veri.: Verifiableness.}
\label{tab:graded_corr}
\vspace*{-0.45cm}
\end{wraptable}

\smallskip\smallskip
\noindent\textbf{Effect on Evaluation Performance.}
Table~\ref{tab:graded_corr} compares Pearson correlations with human judgments under binary vs.\ 3-level alignment, revealing an asymmetry between task types. For close-ended queries, the 3-level variant yields consistent improvements, notably in Precision ($+$0.12) and Verifiableness ($+$0.07), where graded scoring adds discriminative power for borderline evidence. For open-ended queries, however, binary alignment proves more effective, suggesting that finer-grained distinctions may not always translate to better alignment with human judgments. These results suggest that the 3-level variant is beneficial for close-ended evaluation, while binary alignment remains preferable for open-ended settings.

\section{Prompts}
\label{sec:prompts}

We design a series of prompts for LLM-based evaluation, organized around two core evaluation axes: coverage and verifiability. Coverage prompts assess whether the retrieved chunks and generated answer sufficiently address the query requirements, while verifiability prompts determine whether answer claims are grounded in the retrieved evidence. All prompts are provided in Tables~\ref{tab:prompt_query}--\ref{tab:prompt_generator}.

\smallskip\smallskip
\noindent\textbf{Query Decomposition.}
Table~\ref{tab:prompt_query} presents the prompt used for decomposing queries into sub-queries. The prompt guides the model to first classify the query into one of three categories: single-focus, multi-aspect, or multi-step reasoning. Based on this classification, the model applies the corresponding decomposition strategy. Single-focus queries remain unchanged, while multi-aspect queries are decomposed into independent perspectives and multi-step reasoning queries are separated into sequential logical steps. The resulting sub-queries are constrained to be mutually exclusive and collectively cover the original query.

\smallskip\smallskip
\noindent\textbf{Answer Decomposition.}
Table~\ref{tab:prompt_atomic} presents the prompt used for decomposing answer into answer claims. The prompt instructs the model to produce self-contained claims that can be understood independently of the original query. For instance, given the query ``What is the capital of France?'' and the answer ``Paris'', the extracted claim would be ``The capital of France is Paris'' rather than the standalone entity. The prompt also prohibits referential expressions such as ``it'', ``this'', or ``the above'' to ensure each claim carries complete context for subsequent process.

\smallskip\smallskip
\noindent\textbf{Sub-query--Chunk Alignment.}
Table~\ref{tab:prompt_sc_coverage} shows the prompt for evaluating whether a retrieved chunk provides sufficient information to fully answer a given sub-query. This alignment is used to compute coverage-aware soft relevance labels for retriever evaluation, capturing partial coverage that binary relevance labels would miss. The model outputs ``Yes'' only when the chunk contains complete information without truncation or missing details.

\smallskip\smallskip
\noindent\textbf{Sub-query--Claim Alignment.}
Table~\ref{tab:prompt_sf_relevance} shows the prompt for determining whether an answer claim provides necessary information for a specific sub-query. This pairwise alignment identifies which portions of the generated answer address which sub-query, forming the basis for evaluating answer completeness and conciseness.

\smallskip\smallskip
\noindent\textbf{Claim--Chunk Alignment.}
Table~\ref{tab:prompt_cf_verification} shows the prompt for verifying whether answer claims are supported by retrieved chunks. Given a chunk and a set of claims, the model identifies which claims can be grounded in the retrieved evidence. This alignment is used to compute verifiableness, distinguishing claims supported by retrieved context from those potentially derived from the model's internal knowledge.

\smallskip\smallskip
\noindent\textbf{Sub-query--Relevant Claims Coverage.}
Table~\ref{tab:prompt_sf_coverage} shows the prompt for evaluating whether a set of relevant claims collectively provides sufficient information to fully answer a sub-query. Unlike the individual sub-query--claim alignment, this prompt assesses the combined coverage of relevant answer claims, determining whether any critical information is missing for answering the sub-query. This coverage check identifies which sub-queries are fully covered by the answer, enabling computation of completeness and conciseness.

\smallskip\smallskip
\noindent\textbf{Query Type Classification.}
Table~\ref{tab:prompt_query_classification} shows the prompt for categorizing queries as close-ended or open-ended. We use this prompt during benchmark construction to filter queries, ensuring that each dataset (\emph{i.e.}, close-ended and open-ended) contains only queries of the corresponding type.

\begin{table*}[!t]
\centering
\scriptsize
\setlength{\tabcolsep}{4.4pt}
\definecolor{avgcol}{RGB}{235, 235, 235}
\newcolumntype{C}{>{\centering\arraybackslash}m{0.055\textwidth}}
\newcolumntype{A}{>{\centering\arraybackslash\columncolor{avgcol}}m{0.055\textwidth}}
\begin{tabular}{cl c CCCC A CCCC A}
\toprule
& & & \multicolumn{5}{c}{\textbf{Close-ended}} & \multicolumn{5}{c}{\textbf{Open-ended}} \\
\cmidrule(lr){4-8} \cmidrule(lr){9-13}
& Model & Metric & NQ & NewsQA & HotpotQA & FinQA & Avg. & PubMed-QA & LoTTE-Sci. & LoTTE-Tech. & ELI5 & Avg. \\
\midrule
\multirow{15}{*}{\rotatebox{90}{\textsc{Open Source}}} 
& \multirow{3}{*}{Qwen3-4B-Inst.} 
& Comp. & 0.755 & 0.692 & 0.748 & 0.132 & 0.581 & 0.672 & 0.781 & 0.680 & 0.798 & 0.733 \\
& & Conc. & 0.808 & 0.752 & 0.853 & 0.152 & 0.641 & 0.739 & 0.661 & 0.605 & 0.660 & 0.666 \\
& & Veri. & 0.571 & 0.754 & 0.654 & 0.401 & 0.595 & 0.700 & 0.892 & 0.859 & 0.570 & 0.755 \\
\cmidrule(lr){2-13}
& \multirow{3}{*}{Qwen3-30B-Inst.} 
& Comp. & 0.755 & 0.723 & 0.742 & 0.142 & 0.591 & 0.773 & 0.757 & 0.742 & 0.801 & 0.768 \\
& & Conc. & 0.851 & 0.756 & 0.824 & 0.205 & 0.659 & 0.794 & 0.666 & 0.652 & 0.728 & 0.710 \\
& & Veri. & 0.569 & 0.783 & 0.644 & 0.580 & 0.644 & 0.781 & 0.899 & 0.865 & 0.552 & 0.774 \\
\cmidrule(lr){2-13}
& \multirow{3}{*}{Gemma-3-4B-IT} 
& Comp. & 0.700 & 0.632 & 0.642 & 0.233 & 0.552 & 0.582 & 0.652 & 0.660 & 0.684 & 0.644 \\
& & Conc. & 0.784 & 0.692 & 0.716 & 0.298 & 0.622 & 0.628 & 0.553 & 0.555 & 0.624 & 0.590 \\
& & Veri. & 0.526 & 0.709 & 0.661 & 0.374 & 0.568 & 0.739 & 0.847 & 0.842 & 0.584 & 0.753 \\
\cmidrule(lr){2-13}
& \multirow{3}{*}{Gemma-3-27B-IT} 
& Comp. & 0.670 & 0.660 & 0.722 & 0.193 & 0.561 & 0.602 & 0.702 & 0.645 & 0.630 & 0.645 \\
& & Conc. & 0.748 & 0.779 & 0.820 & 0.232 & 0.645 & 0.632 & 0.608 & 0.567 & 0.596 & 0.601 \\
& & Veri. & 0.682 & 0.796 & 0.678 & 0.581 & 0.684 & 0.853 & 0.912 & 0.871 & 0.602 & 0.809 \\
\cmidrule(lr){2-13}
& \multirow{3}{*}{GPT-oss-20B} 
& Comp. & 0.745 & 0.692 & 0.742 & 0.148 & 0.582 & 0.764 & 0.818 & 0.713 & 0.855 & 0.788 \\
& & Conc. & 0.865 & 0.804 & 0.848 & 0.220 & 0.684 & 0.778 & 0.709 & 0.622 & 0.727 & 0.709 \\
& & Veri. & 0.534 & 0.815 & 0.656 & 0.470 & 0.619 & 0.710 & 0.796 & 0.737 & 0.441 & 0.671 \\
\midrule
\multirow{9}{*}{\rotatebox{90}{\textsc{Proprietary}}} 
& \multirow{3}{*}{GPT-5} 
& Comp. & 0.755 & 0.687 & 0.758 & 0.163 & 0.591 & 0.751 & 0.817 & 0.758 & 0.729 & 0.764 \\
& & Conc. & 0.862 & 0.818 & 0.857 & 0.239 & 0.694 & 0.628 & 0.542 & 0.513 & 0.522 & 0.551 \\
& & Veri. & 0.600 & 0.787 & 0.690 & 0.599 & 0.669 & 0.817 & 0.883 & 0.893 & 0.669 & 0.815 \\
\cmidrule(lr){2-13}
& \multirow{3}{*}{Gemini-2.5-pro} 
& Comp. & 0.575 & 0.637 & 0.661 & 0.343 & 0.554 & 0.676 & 0.762 & 0.682 & 0.564 & 0.671 \\
& & Conc. & 0.582 & 0.700 & 0.683 & 0.313 & 0.570 & 0.610 & 0.640 & 0.555 & 0.500 & 0.576 \\
& & Veri. & 0.706 & 0.840 & 0.739 & 0.632 & 0.729 & 0.894 & 0.951 & 0.939 & 0.687 & 0.868 \\
\cmidrule(lr){2-13}
& \multirow{3}{*}{Claude-Sonnet-4} 
& Comp. & 0.660 & 0.697 & 0.678 & 0.403 & 0.609 & 0.782 & 0.834 & 0.750 & 0.812 & 0.795 \\
& & Conc. & 0.597 & 0.683 & 0.755 & 0.331 & 0.591 & 0.675 & 0.660 & 0.575 & 0.641 & 0.638 \\
& & Veri. & 0.731 & 0.925 & 0.702 & 0.721 & 0.770 & 0.825 & 0.887 & 0.877 & 0.610 & 0.800 \\
\midrule
\rowcolor{avgcol}
 & \multirow{3}{*}{\textbf{Avg.}} & Comp. & 0.702 & 0.677 & 0.712 & 0.220 & 0.578 & 0.700 & 0.765 & 0.704 & 0.734 & 0.726 \\
\rowcolor{avgcol}
 & & Conc. & 0.762 & 0.748 & 0.794 & 0.249 & 0.638 & 0.685 & 0.630 & 0.580 & 0.625 & 0.630 \\
\rowcolor{avgcol}
 & & Veri. & 0.615 & 0.801 & 0.678 & 0.545 & 0.660 & 0.790 & 0.883 & 0.860 & 0.589 & 0.781 \\
\bottomrule
\end{tabular}
\vspace{-0.15cm}
\caption{Dataset-level breakdown of generator benchmarking results from Figure~\ref{fig:generator_eval}. We benchmark 8 LLMs spanning open-source and proprietary models using \algname{} with Qwen3-30B-A3B-Inst. as the backbone. Comp.: Completeness, Conc.: Conciseness, Veri.: Verifiableness.}
\label{tab:full-results}
\end{table*}

\begin{table*}[t]
\centering
\footnotesize
\begin{tabular}{@{} p{0.95\linewidth} @{}}
\toprule
\multicolumn{1}{c}{Query Decomposition Prompt} \\
\midrule
\textbf{System:} You are a query analyzer. Decompose complex queries into minimal, essential subqueries.\\\\

\textbf{User:} Decompose the query into independent Core subqueries.\\\\

\textbf{Step 1: Identify Query Type}\\
- Multi-step reasoning: Requires sequential logical steps\\
- Multi-aspect: Requires multiple independent dimensions\\
- Single-focus: One coherent question\\

\textbf{Step 2: Apply Strategy}\\
A) Multi-step reasoning $\rightarrow$ Break into sequential logical steps\\
B) Multi-aspect $\rightarrow$ Break into minimal, non-redundant perspectives\\
C) Single-focus $\rightarrow$ Return the original query EXACTLY as is \\\\

\textbf{Rules:}\\
1. Each subquery must be self-contained\\
2. Do not include background knowledge or assumptions\\
3. Subqueries must be distinct (no overlap)\\
4. Core subquery1 $\cap$ Core subquery2 $\cap$ ... = $\emptyset$\\
5. Core subquery1 $\cup$ Core subquery2 $\cup$ ... = QUERY
\\\\
\textbf{Query:} \textcolor{blue}{\{query\}}
\\
\textbf{Output JSON only:} \\
\{``query\_decomposition'':\\
\quad\{``Core subquery1'': ``...''\}, \\
\quad\{``Core subquery2'': ``...''\}, ...\} \\
\bottomrule
\end{tabular}
\vspace{-0.15cm}
\caption{Prompt for query decomposition.}
\label{tab:prompt_query}
\end{table*}

\begin{table*}[t]
\centering
\footnotesize
\begin{tabular}{@{} p{0.95\linewidth} @{}}
\toprule
\multicolumn{1}{c}{Atomic Fact Extraction Prompt} \\
\midrule
\textbf{System:} You are an expert answer analyzer. Extract atomic facts that are granular but not over-segmented.\\\\

\textbf{User:} Decompose the answer into atomic facts.\\\\

\textbf{Atomic Fact Definition:}\\
- One verifiable fact that can stand alone\\
- Represents a single distinct piece of information\\
- Self-contained with full context (no referential terms like ``that'', ``this'', ``the above'')\\
- Must be understandable WITHOUT seeing the original query

\\\\
\textbf{Decomposition Rules:}\\
- Do NOT extract generic statements like ``The answer is yes/no'' - incorporate the actual claim\\
- If the answer is very short, convert it into a complete statement using context from the query\\
- Keep related information together: if same subject/predicate applies to multiple objects, list them in one fact\\
- Avoid redundant or overlapping facts\\
- Do not overgeneralize beyond what is explicitly stated\\
- ALWAYS produce at least one atomic fact - never return empty

\\\\
\textbf{Critical:} Extract ONLY information explicitly stated in the ANSWER. Do NOT add any background knowledge. Do NOT infer or assume ANY information not present in the answer.

\\\\
\textbf{Query:} \textcolor{blue}{\{query\}} \\
\textbf{Answer:} \textcolor{blue}{\{answer\}}

\\\\
Output as a numbered list (no other text):\\
1. [first atomic fact]\\
2. [second atomic fact]\\
...\\
\bottomrule
\end{tabular}
\vspace{-0.15cm}
\caption{Prompt for atomic fact extraction.}
\label{tab:prompt_atomic}
\end{table*}

\begin{table*}[t]
\centering
\footnotesize
\begin{tabular}{@{} p{0.95\linewidth} @{}}
\toprule
\multicolumn{1}{c}{Sub-query -- Chunk Alignment Prompt} \\
\midrule
\textbf{System:} You are an expert evaluator specializing in Natural Language Inference (NLI) for question-answering systems.\\\\

\textbf{User:} Does the given CHUNK provide COMPLETE information to FULLY answer the SUBQUERY?\\\\

\textbf{Chunk:} \textcolor{blue}{\{chunk\}}\\
\textbf{Sub Query:} \textcolor{blue}{\{subquery\}}
\\\\
\textbf{Rules:}\\
- Answer `Yes' ONLY if you can provide a complete, specific answer\\
- Answer `No' if sentences are cut-off or critical details are missing
\\\\
Only output `Yes' or `No' without any explanation. \\
\bottomrule
\end{tabular}
\vspace{-0.15cm}
\caption{Prompt for sub-query to chunk coverage evaluation.}
\label{tab:prompt_sc_coverage}
\end{table*}

\begin{table*}[t]
\centering
\footnotesize
\begin{tabular}{@{} p{0.95\linewidth} @{}}
\toprule
\multicolumn{1}{c}{Sub-query -- Claim Alignment Prompt} \\
\midrule
\textbf{System:} You are an expert evaluator specializing in Natural Language Inference (NLI) for question-answering systems.\\\\

\textbf{User:} Does the given ATOMIC FACT provide NECESSARY information to answer the SUB QUERY?\\\\

\textbf{Sub Query:} \textcolor{blue}{\{subquery\}} \\
\textbf{Atomic Fact:} \textcolor{blue}{\{fact\}}

\\\\
\textbf{Rules:}\\
- Answer `Yes' ONLY if the fact is directly needed to answer the SUB QUERY\\
- Answer `No' if the fact is not required to answer the SUB QUERY
\\\\
Only output `Yes' or `No' without any explanation. \\
\bottomrule
\end{tabular}
\vspace{-0.15cm}
\caption{Prompt for sub-query to atomic fact relevance evaluation.}
\label{tab:prompt_sf_relevance}
\end{table*}

\begin{table*}[t]
\centering
\footnotesize
\begin{tabular}{@{} p{0.95\linewidth} @{}}
\toprule
\multicolumn{1}{c}{Chunk -- Claim Alignment Prompt} \\
\midrule
\textbf{System:} You are an expert evaluator specializing in Natural Language Inference (NLI) for question-answering systems.\\\\

\textbf{User:} Identify which ATOMIC FACTS can be verified from the CHUNK.\\\\

\textbf{Chunk:} \textcolor{blue}{\{chunk\}} \\
\textbf{Atomic Facts:} \textcolor{blue}{\{atomic\_facts\}}

\\\\

\textbf{Rules:}\\
- Include a fact if it is supported by information in the CHUNK\\
- The chunk should provide evidence for the Atomic fact's validity\\
- Each Atomic fact is evaluated independently
\\\\
\textbf{Output JSON only:} \\
\{``verified\_facts'': \\
\quad [``Atomic fact1'', ``Atomic fact2'', ...] or []\\
\} \\
\bottomrule
\end{tabular}
\vspace{-0.15cm}
\caption{Prompt for chunk to atomic fact verification.}
\label{tab:prompt_cf_verification}
\end{table*}

\begin{table*}[!t]
\centering
\footnotesize
\begin{tabular}{@{} p{0.95\linewidth} @{}}
\toprule
\multicolumn{1}{c}{Sub-query -- Relevant Claims Coverage Prompt} \\
\midrule
\textbf{System:} You are an expert evaluator specializing in Natural Language Inference (NLI) for question-answering systems.\\\\

\textbf{User:} Does the combination of ATOMIC FACTS cover the information needed to answer the SUBQUERY?\\\\

\textbf{Atomic Facts:} \textcolor{blue}{\{facts\}} \\
\textbf{Sub Query:} \textcolor{blue}{\{subquery\}}

\\\\

\textbf{Rules:}\\
- Answer `Yes' ONLY if the ATOMIC FACTS contain information needed for a complete, specific answer\\
- Answer `No' if ANY critical information is missing or the ATOMIC FACTS are insufficient to answer the SUBQUERY
\\\\
Only output `Yes' or `No' without any explanation. \\
\bottomrule
\end{tabular}
\vspace{-0.15cm}
\caption{Prompt for sub-query to atomic facts coverage evaluation.}
\label{tab:prompt_sf_coverage}
\end{table*}

\begin{table*}[!t]
\centering
\footnotesize
\begin{tabular}{@{} p{0.95\linewidth} @{}}
\toprule
\multicolumn{1}{c}{Query Type Classification Prompt} \\
\midrule
\textbf{System:} You are a helpful assistant.\\\\

\textbf{User:} You are a classifier. Decide whether the given query is open-ended or close-ended.\\\\

\textbf{Definitions:}\\
- Close-ended: Questions with specific, factual, or definitive answers (who, what, when, where facts, yes/no questions)\\
- Open-ended: Questions that are subjective, exploratory, or have multiple valid answers
\\\\

\textbf{Query:} \textcolor{blue}{\{query\}}

\\\\
\textbf{Output only in JSON format as:} \\\\
\{``label'': ``open-ended''\} or \{``label'': ``close-ended''\} \\
\bottomrule
\end{tabular}
\vspace{-0.15cm}
\caption{Prompt for query type classification.}
\label{tab:prompt_query_classification}
\end{table*}

\begin{table*}[!t]
\centering
\footnotesize
\begin{tabular}{@{} p{0.95\linewidth} @{}}
\toprule
\multicolumn{1}{c}{Answer Generation Prompt} \\
\midrule
\textbf{User:} Answer the QUERY below. Refer to the provided CONTEXT if needed.
\\\\
\textbf{Query:} \textcolor{blue}{\{query\}} \\
\textbf{Context:} \textcolor{blue}{\{retrieved\_documents\}}
\\\\
\textbf{Instructions:}\\
- Answer only the query directly and naturally\\
- Do not evaluate or analyze the context\\
- Do not mention the context, documents, or sources in your response\\
- Response must be in JSON format only
\\\\
\textbf{Output JSON only:}\\
\{``Answer'': ``your direct answer here''\} \\
\bottomrule
\end{tabular}
\vspace{-0.15cm}
\caption{Prompt for answer generation by the LLM generator.}
\label{tab:prompt_generator}
\end{table*}

\end{document}